\documentclass[letterpaper]{article} 
\usepackage{aaai2026}  
\usepackage{times}  
\usepackage{helvet}  
\usepackage{courier}  
\usepackage[hyphens]{url}  
\usepackage{graphicx} 
\usepackage{natbib}  
\usepackage{caption} 
\usepackage{algorithm}
\usepackage{algorithmic}

\usepackage{newfloat}
\usepackage{listings}
\DeclareCaptionStyle{ruled}{labelfont=normalfont,labelsep=colon,strut=off} 
\floatstyle{ruled}
\newfloat{listing}{tb}{lst}{}
\floatname{listing}{Listing}
\usepackage[utf8]{inputenc}
\usepackage{cuted}
\usepackage{xcolor}         
\usepackage{amsmath}
\usepackage{amssymb}
\usepackage{booktabs}
\usepackage{comment}
\usepackage{adjustbox}
\usepackage{xspace}
\usepackage{multirow}
\usepackage{array}
\usepackage{colortbl} 
\usepackage{cancel}
\usepackage{subfig}

\usepackage[capitalize]{cleveref}
\crefname{section}{Sec.}{Secs.}
\Crefname{section}{Section}{Sections}
\crefname{subsection}{Sec.}{Secs.}
\Crefname{subsection}{Section}{Sections}
\Crefname{table}{Table}{Tables}
\crefname{table}{Tab.}{Tabs.}
\Crefname{definition}{Def.}{Defs}
\Crefname{equation}{Eq.}{Eqs.}
\Crefname{proposition}{Prop.}{Props.}

\newcommand{\problem}{\textcolor{black}{spatial inconsistency}}
\newcommand{\model}{\textcolor{black}{I-LDM}}

\definecolor{yellow_1}{RGB}{255,213,151}
\definecolor{yellow_2}{RGB}{255,225,0}
\definecolor{green_1}{RGB}{98,254,0}
\definecolor{TableGray}{gray}{0.95}
\definecolor{TableBlue}{RGB}{230, 230, 255}
\newcolumntype{g}{>{\columncolor{TableGray}}c}
\newcolumntype{b}{>{\columncolor{TableBlue}}c}

\DeclareMathOperator*{\argmax}{arg\,max}

\newcommand{\red}[1]{{\color{red}#1}}

\title{Towards Spatially Consistent Image Generation: On Incorporating \\ Intrinsic Scene Properties into Diffusion Models}
\author {
    Hyundo Lee\textsuperscript{\rm 1},
    Suhyung Choi\textsuperscript{\rm 1},
    Inwoo Hwang\textsuperscript{\rm 2\textdagger},
    Byoung-Tak Zhang\textsuperscript{\rm 1\textdagger}
}
\affiliations {
    \textsuperscript{\rm 1}AI Institute, Seoul National University\\
    \textsuperscript{\rm 2}Columbia University\\
    \{hdlee, schoi\}@bi.snu.ac.kr, inwoo.hwang@columbia.edu, btzhang@bi.snu.ac.kr
}

\begin{document}

\maketitle

\begin{strip}\centering
\vspace*{-4em}
\includegraphics[width=1.0\textwidth]{AnonymousSubmission/figures/teaser.pdf}
\captionof{figure}{
By co-generating images and aligned intrinsic scene properties, we aim to address the problem of \textit{\problem{}} prevalent in existing text-to-image models.
\textbf{\textcolor{gray}{(Gray)}} Paradoxical image generated by Stable Diffusion 2.1, including inconsistent wall, object (green circle), and floor (orange circle).
\textbf{\textcolor{yellow_1}{(Beige)}}
Our approach generates an image and intrinsic scene properties representing the scene from diverse perspectives, thereby producing a more natural and realistic image.
}
\label{fig:teaser}
\end{strip}

\begin{abstract}
Image generation models trained on large datasets can synthesize high-quality images but often produce spatially inconsistent and distorted images due to limited information about the underlying structures and spatial layouts.
In this work, we leverage intrinsic scene properties (e.g., depth, segmentation maps)
that provide rich information about the underlying scene,
unlike prior approaches that solely rely on image-text pairs or use intrinsics as conditional inputs.
Our approach aims to co-generate both images and their corresponding intrinsics, enabling the model to implicitly capture the underlying scene structure and generate more spatially consistent and realistic images.
Specifically, we first extract rich intrinsic scene properties from a large image dataset with pre-trained estimators, eliminating the need for additional scene information or explicit 3D representations.
We then aggregate various intrinsic scene properties into a single latent variable using an autoencoder.
Building upon pre-trained large-scale Latent Diffusion Models (LDMs), our method simultaneously denoises the image and intrinsic domains
by carefully sharing mutual information so that the image and intrinsic reflect each other without degrading image quality.
Experimental results demonstrate that our method corrects spatial inconsistencies and produces a more natural layout of scenes while maintaining the fidelity and textual alignment of the base model (e.g., Stable Diffusion).
\end{abstract}

\section{Introduction}
\label{sec:intro}
Recent text-to-image (T2I) generation models, notably diffusion models~\cite{ddpm, diffusion_beat_gan, imagen, ldm}, have shown remarkable success in producing diverse and realistic images.
However, these models often produce images with \textit{spatial inconsistencies}\textemdash such as distorted object geometry or implausible scene layouts as in \cref{fig:teaser}\textemdash due to their reliance on pixel-level supervision and lack of structured scene understanding~\cite{sdxl}.
To overcome this, richer and more structured representations beyond pixels are needed to guide the generation process to generate more spatially coherent images.

A natural remedy is to incorporate explicit 3D scene information (e.g. point clouds, meshes) into the generative process~\cite{point_diffusion, shap_e}. 
While effective in grounding geometry, the high cost of data acquisition and rendering limits its scalability and diversity.
Additionally, the reliance on explicit 3D structure makes them ill-suited for abstract, stylized, or artistic scenes.
An alternative approach is to leverage intrinsic scene properties (or simply \textit{intrinsics})—such as depth, surface normals, or segmentation maps—which offer structured, complementary views of the same scene. 
Here, we use the term \textit{intrinsics} in the sense of \citet{intrinsic}, referring to image-inherent scene attributes rather than camera intrinsics in multi-view geometry.
However, prior work leveraging intrinsics has not aimed to improve spatial consistency, but rather focused on conditional generation given external intrinsic inputs~\cite{controlnet}. 
This reliance on additional signals limits their applicability to general T2I generation and fails to model the underlying scene structure.

To address this issue, we propose a new approach that improves spatial consistency by jointly modeling both images and their intrinsic scene properties.
As illustrated in \cref{fig:teaser}, intrinsics provide complementary geometric and semantic cues that help delineate object boundaries and spatial layouts, especially in perceptually ambiguous regions.
By learning a joint distribution over images and intrinsics, our model captures multiple structured views of a scene, thereby significantly reducing spatial inconsistencies in the generation process.
In contrast to conditional approaches that model a unidirectional relationship, our joint modeling allows intrinsics and images to regularize each other during training.
Incorporating intrinsics also offers practical advantages: they can be efficiently extracted from large-scale 2D datasets using pre-trained models, without the need for 3D data or manual labeling.
Moreover, since large T2I models implicitly contain intrinsic knowledge~\cite{intrinsic_lora, intrinsic_knowledge, intrinsic_knowledge2}, we leverage this capacity with only minimal adjustments, without compromising the quality and capabilities of the original model.

To this end, we propose Intrinsic Latent Diffusion Models (\model{}), which extend existing T2I models~\cite{ldm, sdxl, pixart} to jointly generate images and their corresponding intrinsics. 
To enable efficient joint generation, we first train an autoencoder that encodes multiple intrinsic channels into a single latent space. 
Since naive fine-tuning might compromise the performance of the base model, we retain the architecture and weights for generating images while adopting LoRA weights~\cite{lora} to generate intrinsics separately.
Furthermore, we introduce a cross-domain weight scheduling mechanism that shares self-attention between the image and intrinsic domains during the denoising process, promoting alignment without introducing visual artifacts.

By learning from both perceptual and structural views of a scene, \model{} produces more faithful and structured images without compromising the visual fidelity of strong base models.
We evaluate our model under general text prompts and challenging multi-object arrangements, showing that \model{} produces clearer separation and more realistic spatial layouts.
We show that \model{} generates more realistic images and achieves higher scores in human preference estimators 
while maintaining the fidelity and text alignment of the base model.
Furthermore, we demonstrate that learning to co-generate intrinsics helps to model complex structures by providing consistent structural guidance in dynamic hand pose generation tasks.

\section{Related work}
\label{sec:related}
\subsection{Large-scale Diffusion Models}
Recently, image synthesis has achieved remarkable results based on diffusion models~\cite{diffusion_flow, sdv3, pixart}, which generate images by denoising random noise~\cite{diffusion2015, diffusion2019, ddpm}. 
Diffusion models have evolved to enable natural language-based control~\cite{imagen, glide, cfg} using language models such as CLIP~\cite{clip}. 
Latent diffusion models (LDMs)~\cite{ldm} apply perceptual compression via VQ-VAE~\cite{vqvae} to learn the denoising process in perceptually compressed latent space, facilitating high-resolution image synthesis. 
Models such as DALL-E~\cite{dalle3} or Stable Diffusion~\cite{ldm, sdxl} have been trained on large-scale text-image pair datasets containing billions of examples, allowing them to generate images conditioned on general natural language without restricting domains~\cite{laion}. 
Our work is based on production-ready Stable Diffusion models, using fine-tuning techniques with LoRA~\cite{lora} to maintain their quality and capability.

\begin{figure*}[t]
  \centering
  \includegraphics[width=0.98\textwidth]{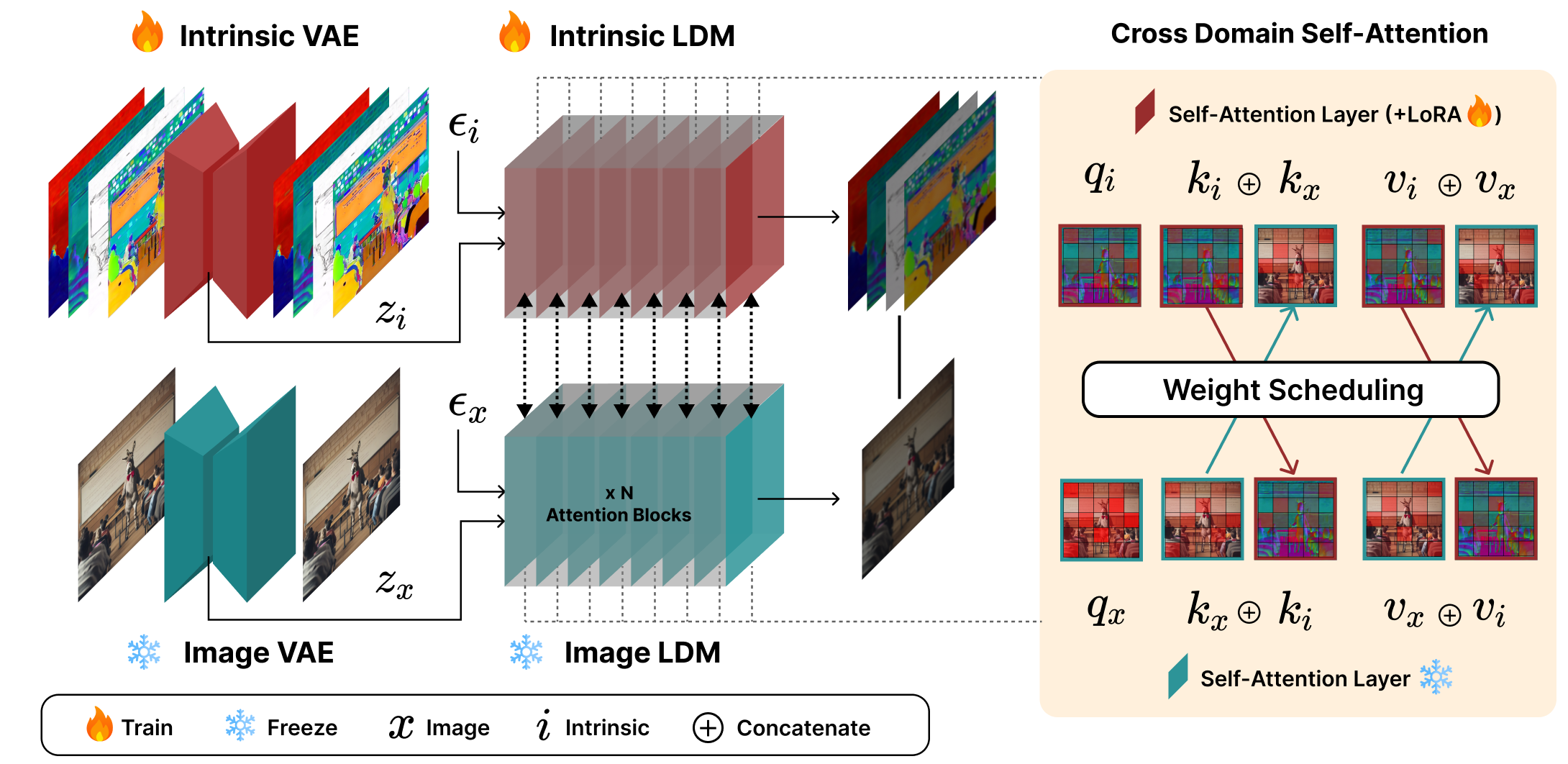}
  \caption{
  The overall architecture of \model{}. 
  \textbf{(Left)} We first train an intrinsic VAE to encode all intrinsics into a single latent variable. 
  \textbf{(Middle)} Then, we train the LoRA weights of the self-attention layers included in the diffusion network of the intrinsic domain to learn the denoising process.
  \textbf{(Right)} We employ a weight scheduling mechanism for exchanging self-attention with the image domain. 
  As a result, \model{} simultaneously generates spatially consistent images and intrinsics during inference.
  }
  \label{fig:architecture}
\end{figure*}

\subsection{Reflecting Physical Structure in LDMs}
Diffusion models trained on pairs of text images struggle to accurately capture the physical structures and spatial layout of scenes, occasionally producing distorted or inconsistent images~\cite{sdxl, spatial_inconsistency}. 
One solution is to incorporate 3D data~\cite{point_diffusion, point_diffusion2, shap_e}, such as point clouds~\cite{shapenet}, CAD models~\cite{shapenet2}, or polygonal meshes~\cite{objaverse, objaverse_xl}.
Despite effectively capturing geometry, this approach is limited to specific objects and struggles with general or abstract text prompts due to the high cost and inflexibility of 3D data.
Instead, we leverage intrinsic scene properties to capture underlying structures while maintaining flexibility for diverse textual inputs.

In image generation, intrinsic scene properties~\cite{intrinsic} have been widely used as conditional controls, including SDEdit~\cite{sdedit}, sketch-guided generation~\cite{sketch_to_image}, and ControlNet~\cite{controlnet}, focusing on aligning outputs to given intrinsic conditions.
In contrast, we aim to co-generate intrinsics alongside images, enabling the model to learn and reflect the distribution of underlying scenes.

Recent studies also explore generating intrinsics from images using diffusion models, including depth maps and surface normals~\cite{geowizard, diffusion_geo1, diffusion_geo2, diffusion_geo3, marigold}, segmentation maps~\cite{diffusion_seg1, diffusion_seg2, diffusion_seg3}, and other scene properties~\cite{intrinsic_lora}. 
However, our goal is not to reconstruct exact geometry, but to guide image generation by jointly modeling intrinsic and image domains, using only text conditions.

A prior approach generates surface normals from noise using GANs and then conditions image synthesis on them~\cite{intrinsic_gan}.
However, generating intrinsics without considering the image domain can lead to spatial inconsistencies, as they lack awareness of the full scene. 
Another related line of work is LDM3D~\cite{ldm3d}, which generates depth maps aligned with images from text.
However, their primary goal is to generate RGB-D images and 360$^\circ$ views.
In contrast, we focus on enhancing spatial consistency by jointly modeling image and intrinsic representations during generation. 
We empirically compare our method with these prior approaches to validate the effectiveness in improving spatial consistency and image quality.

\section{Method}
\label{sec:method}

In this section, we utilize intrinsics to reflect the underlying scene in generated images more faithfully.
The theoretical motivation is discussed in \cref{sec:method-formulation}.
We first present our model architecture, which incorporates cross-domain self-attention and a weight scheduling mechanism to co-generate images and aligned intrinsics without compromising image quality.
We then introduce a practical approach for handling multiple intrinsics simultaneously.

\begin{figure*}[ht]
  \centering
  \includegraphics[width=0.99\textwidth]{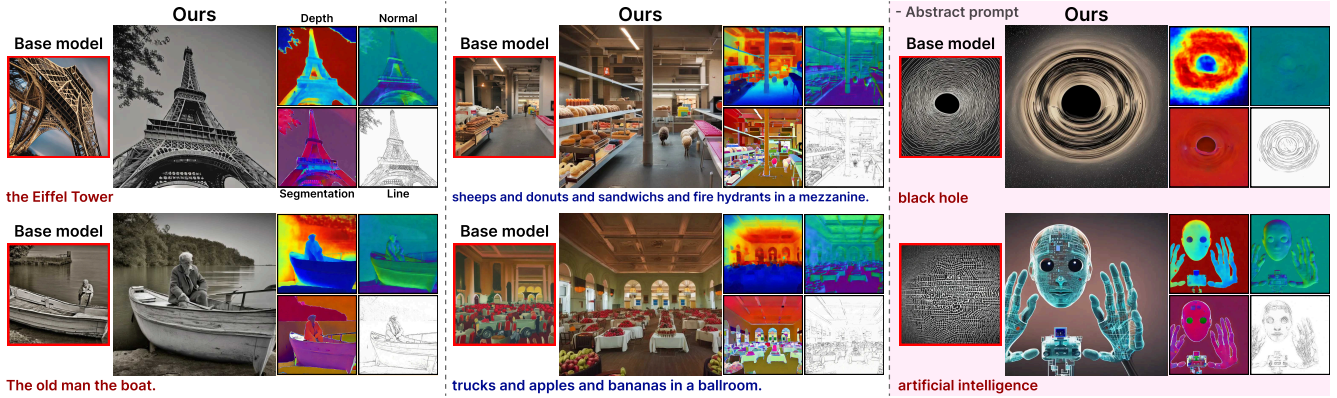}
  \caption{
  Qualitative analysis of generated images and co-generated intrinsic scene properties from \model{} (clockwise from top left: depth map, surface normal, line drawing, and segmentation map). The red boxes indicate generated images of the base model. The top and bottom rows visualize samples with Drop and Gaussian weight scheduling, respectively. Red and blue captions denote samples from the Parti and Multi prompts, respectively.
  }
  \label{fig:qualitative}
\end{figure*}

\subsection{Intrinsic Latent Diffusion Model}
\label{sec:method-ildm}
In this work, we consider four intrinsic scene properties: depth map, surface normal, segmentation map, and line drawing.
A straightforward way to generate an image-intrinsic pair $(x, i)$ given a text condition $c$ is to augment pre-trained T2I models with intrinsic features\textemdash either by adding extra channels or using spatial addition\textemdash and fine-tune them.
However, we found that this significantly harms the image quality, as demonstrated in \cref{subsec:exp_general}.

Instead, we propose a method using a large LDM model consisting of self-attention layers (e.g., Stable Diffusion~\cite{ldm, sdxl}, PixArt~\cite{pixart}) as a base model that fully exploits its capabilities without compromising its quality.
As depicted in \cref{fig:architecture}, we devise separate models for the image domain and the intrinsic domain, preserving the architecture and weights the same as the base model for the image domain. 
To train the intrinsic domain, we adapt LoRA~\cite{lora} to fine-tune only the attention layers, leveraging the inherent knowledge of the intrinsics~\cite{intrinsic_lora, intrinsic_knowledge}.

Second, we employ cross-domain self-attention~\cite{geowizard} to align the spatial arrangements between the two domains and allow them to reflect each other's information. 
Specifically, we concatenate the keys and values from the self-attention layers at the same stage as follows:
\begin{align}\label{eq:cross-domain-self-attention}
\begin{split}
    q_x &= Q_x\cdot z_x,\quad q_i = Q_i\cdot z_i, \\
    k_x &= k_i = (K_x \cdot z_x) \oplus (K_i \cdot z_i), \\
    v_x &= v_i = (V_x \cdot z_x) \oplus (V_i \cdot z_i), \\
\end{split}
\end{align}
where the subscript $x, i$ denotes each domain,
$z$ denotes intermediate representations, 
$\oplus$ denotes concatenation,
$q, k, v$ denotes the query, key, and value, and
$Q, K, V$ denotes the corresponding weight matrix. 

Importantly, we introduce a weight scheduling mechanism for cross-domain self-attention, i.e., $w^{(l, t)}$, in contrast to prior work without weighting~\cite{geowizard}.
This is because we found that the received information from the intrinsic model during image sampling often incurs undesirable artifacts in the generated image, leading to a significant degradation in image quality. 
Therefore, we introduce the scheduling of the weight of the exchanged key and values across domains for each block index $l$ and each timestep $t$.
Specifically, we fix the weight to 1 during training, while introducing two different weight scheduling strategies during sampling, Drop and Gaussian, as follows:
\begin{align} \label{eq:schedule}
\text{(Drop)} \quad &w_{d}^{(l, t)} = 
\begin{cases}
1 & \text{if} \quad (l \in L) \wedge (t \leq \tau), \\
0 & \text{otherwise}, \\
\end{cases} \\
\text{(Gaussian)} \quad &w_{g}^{(l, t)} = \alpha \cdot \exp(-(t-\tau)^2 / \sigma^2),
\end{align}
where $L, \tau, \sigma, \alpha$ are hyperparameters.
We then change the self-attention process as follows:
\begin{align}\label{eq:attention}
\begin{split}
    Attn_x^{(l, t)} &= \textit{softmax}\left(\frac{q_x \cdot k_x^T}{\sqrt{d_k}}+ \log\left(W^{(l, t)}\right)\right)v_x, \\
    Attn_i &= \textit{softmax}\left((q_i \cdot k_i^T)/\sqrt{d_k}\right)v_i, \\
\end{split}
\end{align}
where $d_k$ denotes the inner dimension of $q, k$,
and $W^{(l, t)}$ denotes a $N\times 2N$ weight matrix 
where the first through $N$ columns are filled with 1, and the $N+1$ through $2N$ columns are filled with $w^{(l, t)}$.
The difference from the original image denoising process is that
the intrinsic model affects the image model with a weight $w^{(l, t)}$ by passing its intermediate key and value vectors.
We assign a low weight to the outer blocks of the diffusion network and for low $t$, where relatively fine-grained detail is denoised.
We also assign low weight to very early steps of the denoising process, which greatly impacts the image.
This adjustment allows us to mitigate the influence of artifacts on images, thereby preserving the quality of the generated images.

For the loss function, we use the conventional DDPM loss to train LoRA weights $\theta'$ in the network $\epsilon_{\theta, \theta'}=\left(\epsilon^{(x)}_{\theta, \theta'}, \epsilon^{(i)}_{\theta, \theta'}\right)$ for the image and intrinsic domains separately, 
with assigning a balancing factor of $\lambda$:
\begin{align}\label{eq:loss}
\begin{split}
    \mathcal{L} = \;\mathbb{E}_{x_0,i_0,c, \epsilon_x, \epsilon_i, t}&[||\epsilon^{(x)}_{\theta, \theta'}(x_t, i_t, t, c) - \epsilon_x||^2_2 \\
    + \lambda \cdot &||\epsilon^{(i)}_{\theta, \theta'}(x_t, i_t, t, c) - \epsilon_i||^2_2],
\end{split}
\end{align}
where $\theta$ is the parameters of the base model, 
and $x_t, i_t$ denotes encoded $x_0, i_0$ with mixed noise $\epsilon_x, \epsilon_i$ respectively.
Thus, we train $\theta'$ not only to generate intrinsics but also to provide useful information to the image by reducing $||\epsilon^{(x)}_{\theta, \theta'} - \epsilon_x||^2_2$.
We independently sample $\epsilon_x$ and $\epsilon_i$ during the training and use the same $\epsilon$ during sampling.

\begin{figure*}[t]
  \centering
  \includegraphics[width=0.94\textwidth]{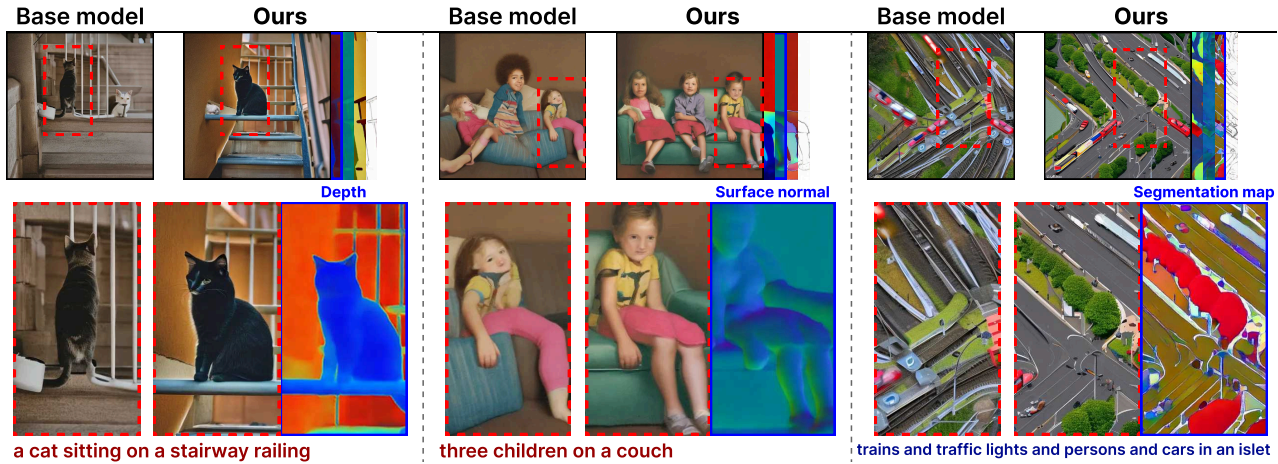}
  \caption{
  Comparison of base and \model{} with generated intrinsics, providing key cues for reducing spatial inconsistencies.
  }
  \label{fig:zoom}
\end{figure*}

\subsection{Intrinsic Space Encoding}
\label{sec:method-ise}
While incorporating various intrinsics enriches scene understanding, increasing their number introduces two main challenges:
(1) higher training and sampling overhead, and (2) learning difficulty due to distributional mismatch between intrinsics and images.
For instance, random noise sampled from a standard normal distribution tends to generate unnatural surface normals, where most pixels are green or blue~\cite{intrinsic_lora, flawed}.
We analyze the latent distribution for images and intrinsics in~\cref{appendix:latent}. 

To address this, we train an intrinsic VAE that encodes all intrinsics into a shared latent variable with the same dimensionality as the base model.
This is achieved by extending the base VAE with additional CNN channels and fine-tuning it.
During training, we randomly zero out individual intrinsics to prevent any particular intrinsic from being encoded dominantly, ensuring that the encoded latent representation reflects information from all intrinsics.
As a result, our method enables efficient two-pass denoising for images and intrinsics, reducing memory from cross-domain attention.
It also aligns intrinsic and image distributions, allowing minimal LoRA weight adjustments and eliminating the need for zero terminal SNR~\cite{flawed}.
Ablations of intrinsic VAE are presented in~\cref{table:ablation}.

\section{Experiments}
\label{sec:exp}

Here, we investigate the effectiveness of intrinsics with \model{} for generating natural and spatially consistent images.
Experimental details are provided in \cref{appendix:exp-details}.

\paragraph{Metrics.}
Our goal is to alleviate spatial inconsistencies by accurately reflecting the underlying scene while maintaining the quality and capability of the base model.
However, precisely quantifying spatial inconsistency is inherently challenging, as ground-truth structural annotations do not exist for synthesized images.
For this, we first utilize metrics based on human preference prediction, including ImageReward (IR)~\cite{imagereward} and Human Preference Score (HPSv2)~\cite{hpsv2}.
To further assess the improvement in spatial inconsistency, we evaluate AI preference using GPT-4o, asking \textit{which image better reflects the actual physical scene} between the base model and ours.
Finally, to evaluate whether our method maintains the capability of the base diffusion model, we employ conventional CMMD~\cite{cmmd}
and CLIP score~\cite{clipscore} to measure fidelity and text alignment, respectively.
We measure the distribution discrepancy for CMMD in \cref{subsec:exp_general} by comparing with images from the base model, as real images are not available in this setting. 
This allows us to assess how well each method preserves the base model’s quality, rather than indicating improvements over it.

\begin{table*}[t]
    \centering
    \caption{
    Quantitative results on generated images for each prompt in Parti and Multi prompts. 
    The evaluation is conducted using images sampled from three seeds for each prompt. 
    The best result is bolded, and the second-best result is underlined. 
    }
    \begin{adjustbox}{width=0.85\textwidth}
    \begin{tabular}{lccggccgg}
    \toprule
                & \multicolumn{4}{c}{\begin{tabular}[c]{@{}l@{}l@{}l@{}}Parti prompt\end{tabular}} & \multicolumn{4}{c}{\begin{tabular}[c]{@{}l@{}l@{}l@{}}Multiple objects\end{tabular}} \\
    \cmidrule(lr){2-5} \cmidrule(lr){6-9}
    Method          & CMMD($\downarrow$)      & CLIP($\uparrow$)      & ImageReward($\uparrow$)   & HPSv2($\uparrow$)     
                    & CMMD($\downarrow$)      & CLIP($\uparrow$)      & ImageReward($\uparrow$)   & HPSv2($\uparrow$)   \\ \midrule  
    Base            & -     & 0.2725    & 0.2998        & 0.2497    
                    & -     & 0.2671    & -0.7389       & 0.2355      \\
    Base $\times$2  & \textbf{0.00488}     & \underline{0.2726}    & 0.3647        & 0.2548      
                    & \textbf{0.00763}     & \underline{0.2676}    & -0.7342       & 0.2395      \\
    Base +CNN       & 1.07265     & 0.2164    & -1.5378        & 0.1713     
                    & 2.51353     & 0.1986    & -1.8801        & 0.1650       \\
    Base +SpatialAdd& 0.39911     & 0.2413    & -0.7374        & 0.2058     
                    & 1.12689     & 0.2214   & -1.5315        & 0.1807       \\                
    LDM3D           & 0.57590     & 0.2693    & 0.1254          & 0.2349      
                    & 1.20401     & 0.2472    & -1.0290      & 0.2096       \\        
    LDM3D+          & 0.11885     & 0.2644    & 0.0499       & 0.2357      
                    & 0.33486     & 0.2575    & -0.6855       & 0.2298       \\
    ControlNet-Norm
                    & 0.04578     & 0.2697    & 0.3444         & 0.2548      
                    & 0.11182     & 0.2599    & -0.6445       & 0.2388       \\
    ControlNet-Seg
                    & 0.08571     & 0.2672    & 0.3144         & 0.2550      
                    & 0.36383     & 0.2521    & -0.6504       & 0.2340       \\
    ControlNet-All
                    & 0.20289     & 0.2671    & 0.2900         & 0.2549      
                    & 0.47350     & 0.2540    & -0.6877       & 0.2342       \\            
    \textbf{\model{}} (Drop)      & 0.02074     & 0.2717    & \underline{0.3843}        & \underline{0.2560} 
                    & \underline{0.03707}     & 0.2664    & \underline{-0.6148}       & \textbf{0.2461}    \\
    \textbf{\model{}} (Gaussian)    & \underline{0.01538}     & \textbf{0.2739}    & \textbf{0.4416}        & \textbf{0.2582}   
                    & 0.08228     & \textbf{0.2687}    & \textbf{-0.5739}       & \underline{0.2450}   \\
    
\bottomrule
\end{tabular}
\end{adjustbox}
\label{table:general}
\end{table*}

\begin{table}[t]
    \centering
    \caption{
    LLM evaluation of spatial consistency between images generated by the base model and ours. Ours, Tie, Base denote GPT-4o preference percentages.
    }
    \begin{adjustbox}{width=\linewidth}
    \begin{tabular}{lcccccc}
    \toprule
    & \multicolumn{3}{c}{\begin{tabular}[c]{@{}l@{}l@{}}Parti prompt\end{tabular}} & \multicolumn{3}{c}{\begin{tabular}[c]{@{}l@{}l@{}}Multiple objects\end{tabular}} \\
    \cmidrule(lr){2-4} \cmidrule(lr){5-7}
    Method  & Ours($\uparrow$)  & Tie       & Base($\downarrow$)    & Ours($\uparrow$)  & Tie       & Base($\downarrow$) \\
    \midrule
    \model{} (D)        & 41.9\%            & 32.8\%    & 25.2\%                                & 47.7\%            & 36.6\%    & 15.7\% \\
    \model{} (G)    & 41.0\%            & 34.9\%    & 24.1\%                                & 44.3\%            & 39.4\%    & 16.3\% \\
\bottomrule
\end{tabular}
\end{adjustbox}
\label{table:gpt}
\end{table}

\paragraph{Baselines.}
We employ Stable Diffusion 2.1 (SD2.1) as a pre-trained base model~\cite{ldm}.
Comparison of its configuration with ours in terms of model size, inference time, and memory usage is in \cref{table:config}.
For baselines, we first implement LDM3D~\cite{ldm3d} based on SD2.1, training a VAE to encode both images and depth into the existing latent space, and then fine-tune UNet.
We also extend this method with four intrinsics that we use, denoted as LDM3D+.
We also consider a naive method that adds extra CNN channels for intrinsics to the UNet and fine-tunes it (+CNN), and a variant that injects intrinsic features via spatial addition using the ControlNet architecture (+SpatialAdd).
To compare with a two-stage generation approach, we first generate single or multiple intrinsics from text using our intrinsic LDM trained without images, and then use them to condition image generation via ControlNet, denoted as ControlNet-(Intrinsic).
Finally, we evaluate base models with doubled sampling steps (denoted as $\times 2$).

\paragraph{Implementation details.}
For the training dataset, we use a subset of 542k images with aesthetic scores of 6.5 or higher
from LAION-5B dataset~\cite{laion} in which the base model was trained.
To estimate intrinsics, we use Metric3D~\cite{metric3d} for depth maps and surface normals, SAM~\cite{sam2} for segmentation maps, and a publicly available image-to-line drawing model.
For training intrinsic VAE, we zero-mask each intrinsic with a probability of 0.1.
To train the denoising process in the intrinsic domain, we adapt LoRA to self-attention weights with rank 32.
The resolution of the images for training is set to 512$\times$512, and 768$\times$768 for sampling.
All images were sampled with the CFG scale 7.5 and the sampling steps 25 (50 steps for $\times 2$).
For the hyperparameters of our model in \cref{eq:schedule}, we set $\tau=900, L=\{3,4,5,6,7\}$ for $w_{d}$, $\alpha=1, \tau=800, \sigma=100$ for $w_{g}$.

\subsection{Learning Intrinsics on T2I Generation} \label{subsec:exp_general}

We perform two tasks to evaluate the quality and spatial consistency of T2I generation with and without learning intrinsics.
First, we assess general text conditions using images generated from Parti prompts (Parti)~\cite{parti}, designed to test capability across various categories and challenges.
However, some prompts are too abstract or simple to evaluate spatial inconsistency (e.g., ``energy" or ``an orange").
Therefore, we design a second task, the multiple objects prompts (Multi), which constructs complex scenes where structural distortion can be more pronounced by placing many objects. 
We randomly sample 3 to 5 classes from the MS-COCO dataset~\cite{coco} and one place from Places365~\cite{places365}, and construct 1,000 prompts in the format \textbf{``\{class1\}s and \{class2\}s and ... in a \{place\}"}.

\cref{table:general,table:gpt,fig:qualitative} summarize both quantitative metrics and qualitative results on the Parti and Multi prompts.
Compared to the base model and other intrinsic-based approaches, learning to co-generate multiple intrinsics leads to better alignment with the physical structure of the scene without compromising image quality (see \cref{appendix:comparison_baseline} for detailed baseline analysis).
\cref{fig:zoom} emphasizes that the base model may misrepresent the physical relationship between each object and incorrectly represent the boundaries of objects.
In contrast, the intrinsics generated by our method clearly distinguish and reflect them in the image, resulting in a more accurate representation.

\begin{table}[t]
    \centering
    \caption{Quantitative results on downscaling the intrinsic resolution provided by the intrinsic estimators for training. 
    }
    \begin{adjustbox}{width=0.95\linewidth}
    \begin{tabular}{lccccc}
        \toprule
        Method & Resolution & CMMD($\downarrow$)      & CLIP($\uparrow$)      & ImageReward($\uparrow$)   & HPSv2($\uparrow$) \\ \midrule 
        \multirow{3}{*}{\model{} (D)}   &  256    & 0.02027     & 0.2732    & 0.4572        & 0.2590   \\
                                &  384    & 0.02122     & 0.2739   & 0.5129        & 0.2606   \\
                       &  \textbf{512}    & 0.02730     & 0.2717   & 0.4719        & 0.2622   \\ \midrule
        \multirow{3}{*}{\model{} (G)}   &  256    & 0.01431     & 0.2742    & 0.4960        & 0.2606   \\
                                &  384    & 0.01466     & 0.2756   & 0.5311        & 0.2618   \\
                       &  \textbf{512}    & 0.01633     & 0.2741   & 0.4803        & 0.2611   \\
        \bottomrule
        \end{tabular}
        \end{adjustbox}
    \label{table:robustness}
\end{table}

While our method reflects valuable information from the intrinsic, we emphasize that it also protects from the effects of ambiguities or inaccuracies from the intrinsic domain.
\paragraph{Preserving diversity.}
Although abstract scenes can be difficult to represent using geometric intrinsics, such as the "black hole" example in \cref{fig:qualitative}, we demonstrate that our approach does not constrain the diversity of the base model.
This is quantitatively supported by maintaining CMMD and CLIP scores across various textual conditions provided by Parti prompts.
Analysis on various categories and challenges is reported in \cref{appendix:diversity}.
This is achieved by disentangling the latent representations of images and intrinsics, enabling diversity of images beyond geometric constraints.

\paragraph{Robustness to inaccurate intrinsics.}
An intriguing property of our framework is that we utilize existing intrinsic estimators without the additional cost of data collection.
To investigate how our model performs given an imperfect intrinsic estimation, we downscale the resolution of the intrinsic for training. \cref{table:robustness} shows that our model is robust to such errors or noises from intrinsic estimators. 
Moreover, \cref{appdix_fig:attention} in \cref{appendix:robustness} shows that meaningful information can still be exchanged despite misalignment across domains. 
This is due to the design of cross-domain self-attention process (\cref{eq:attention}), which estimates cross-domain attention within the entire set of patches rather than just the same region.

\begin{figure}[t!]
  \centering
  \includegraphics[width=0.99\linewidth]{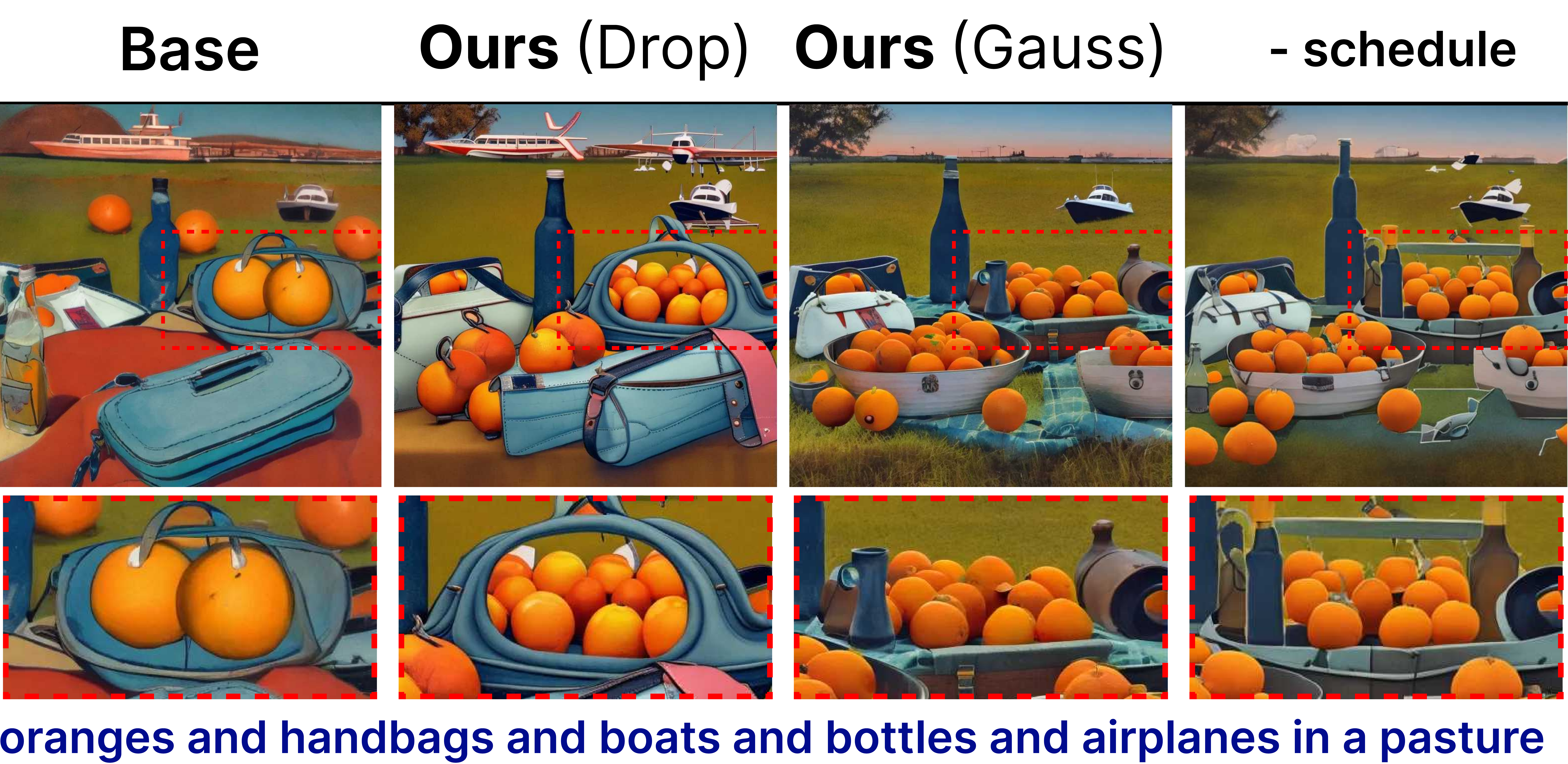}
  \caption{
  Comparison of the base model and \model{} with Drop, Gaussian, and no weight scheduling.
  }
  \label{fig:schedule}
\end{figure}

\paragraph{Drop vs. Gaussian.}
Analysis of different weight scheduling are visualized in \cref{fig:schedule}.
With drop scheduling, there is a tendency to improve fine-grained details while maintaining the overall composition of the image. 
In contrast, Gaussian scheduling typically adjusts the overall image composition.
On the other hand, completely removing the weight scheduling from the inference process degrades the quality of the images.
However, fine-tuning approaches (+CNN, +SpatialAdd, LDM3D) suffer from severe image degradation and do not provide meaningful improvements in spatial inconsistency.
See \cref{appendix:qualitative} for further analysis.

\subsection{Complex Hand Structure Generation}

The human hand, with its intricate articulations that result in highly complex structures and self-occlusions, is a well-known example of a particularly challenging object to synthesize.
We investigate how learning intrinsics affect the generation of such complex structures.
First, since the base model tends to generate hand images with limited poses, we generate diverse and complex poses by training a dataset with various hand poses.
Specifically, we extract 3,493 distinct samples from InterHand dataset~\cite{hand}. 
We then train the base model by applying LoRA
without any annotations or additional information.
The prompt is fixed as ``A close-up of a hand with a black background".
We use a hand detection model~\cite{hand_detection} to measure the detection rate and confidence scores for detected cases, assessing whether generated hands follow an accurate structure.

We applied hand LoRA trained on the base model to the image domain of \model{} to investigate how learned intrinsic information affects the generation of hands with dynamic poses. 
\cref{table:hand} and visualizations in \cref{appendix:hand} summarize the experimental results, which show that reflecting intrinsics substantially improves the generation of accurate hand structures.
This experiment supports our claim that incorporating intrinsics helps generate complex structures.

\begin{table}[t]
    \centering
    \caption{
    Quantitative results on 1000 generated images trained with InterHand samples. 
    The best result is bolded.
    }
    \begin{adjustbox}{width=\linewidth}
    \begin{tabular}{lccgg}
    \toprule
    Method          & CMMD($\downarrow$)      & HPSv2($\uparrow$)     & Detection($\uparrow$) & Confidence($\uparrow$)  \\ \midrule  
    Base~\cite{ldm}           & 0.7222    & 0.1475    & 29.82\%   & 0.9310     \\
    Base $\times$2 & \textbf{0.6906}    & 0.1472    & 30.22\%   & 0.9271     \\
    \textbf{\model{}} (Drop)     & 0.7750    & 0.1487    & 28.20\%   & 0.9242     \\
    \textbf{\model{}} (Gauss)    & 0.7051    & \textbf{0.1506}    & \textbf{39.40\%}   & \textbf{0.9384}     \\
    
\bottomrule
\end{tabular}
\end{adjustbox}
\label{table:hand}
\end{table}

\begin{table}[t]
    \centering
    \caption{Results on ablating each intrinsic, model settings.}
    \begin{adjustbox}{width=0.9\linewidth}
    \begin{tabular}{lcccc}
        \toprule
                    & CMMD($\downarrow$)      & CLIP($\uparrow$)      & ImageReward($\uparrow$)   & HPSv2($\uparrow$) \\ \midrule 
        \textbf{\model{}}  & 0.01633     & 0.2741    & 0.4803        & 0.2611    \\
        - Depth     & 0.01144     & 0.2720    & 0.3914        & 0.2578    \\
        - Normal    & 0.01299     & 0.2742    & 0.4655        & 0.2592    \\
        - Line      & 0.01919     & 0.2746    & 0.5074        & 0.2615    \\
        - Segment   & 0.01872     & 0.2727    & 0.4302        & 0.2590    \\
        \midrule
        -$w^{(l, t)}$& 0.05460     & 0.2706    & 0.3852        & 0.2556    \\
        -VAE        & 0.01836     & 0.2702    & 0.3545        & 0.2551    \\
        LoRA-only   & 0.03314     & 0.2747    & 0.3617        & 0.2493    \\
        +resolution & 0.08452     & 0.2640    & 0.1240        & 0.2470    \\
        +$w_{train}$ & 0.03827     & 0.2716    & 0.3920        & 0.2552    \\
        \bottomrule
        \end{tabular}
        \end{adjustbox}
    \label{table:ablation}
\end{table}

\subsection{Adapting \model{} to Other Base Models}
\begin{figure}[t!]
  \centering
  \includegraphics[width=0.95\linewidth]{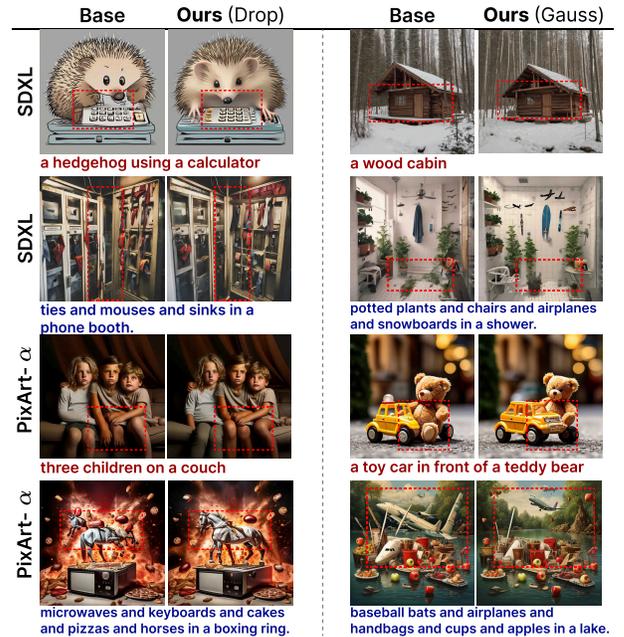}
  \caption{
  Examples of generated images of \model{} using SDXL and PixArt-$\alpha$ as base models.
  }
  \label{fig:others}
\end{figure}

To verify the applicability of our method with more scalable or different architectures, we additionally implement \model{} based on SDXL~\cite{sdxl} with 2.5B parameters and based on PixArt-$\alpha$~\cite{pixart} with DiT architecture~\cite{dit}.
\cref{fig:others} visualizes the qualitative results of our model. 
Samples with SDXL demonstrate that even with a more scalable base model, there are still spatial inconsistencies that \model{} alleviates.
The example of PixArt-$\alpha$ also shows that our method is generally applicable to Transformer-based architectures.
For quantitative analysis and more qualitative results, see \cref{appendix:quantitative,appendix:qualitative}.

\subsection{Ablation Study and Analysis} \label{subsec:ablation}

We perform ablation studies on each intrinsic and the training and sampling settings. 
Evaluations are performed on Parti prompts, using $w_g^{(l, t)}$ unless specifically noted.
More detailed analysis including visualizations are provided in \cref{appendix:abl_intrinsic} and \cref{appendix:abl_settings}.

\paragraph{Effects of each intrinsic scene properties.}
First, we perform an ablation study on each intrinsic scene property that composes \model{}.
\cref{table:ablation} summarizes the effect of ablating each intrinsic, indicating that learning for each intrinsic shows meaningful improvement. 
While line drawing has a small negative effect on metrics, we qualitatively observe that it enhances the generation of object shapes. 

\paragraph{Ablations on training settings.}
Next, we conduct an ablation study on the training and sampling process.
As shown in \cref{table:ablation}, removing weight scheduling causes noticeable artifacts from the intrinsic domain, and excluding the intrinsic VAE results in inconsistent intrinsic generation.
Training LoRA weights without image domain degrades intrinsic quality due to lack of image domain information.
Additionally, using scheduling weights or increasing the resolution to 768×768 during training negatively impacts image quality.

\section{Conclusion}
\label{sec:conclusion}
To mitigate spatial inconsistencies in image generation, we proposed learning intrinsic scene properties aligned with the images.
We presented \model{}, a model that jointly generates images and intrinsics (i.e. depth map, surface normal, line drawing and segmentation map).
We carefully designed our architecture to share valuable self-attention information between the image and the intrinsics while preserving the quality and capabilities of the base model.
Experimental results on Parti, Multi prompts, and InterHand demonstrated the effectiveness of our approach.

\bibliography{main}

\appendix
\onecolumn
\setcounter{page}{1}

\begin{center}
  {\Large\textbf{Towards Spatially Consistent Image Generation: On Incorporating \\ Intrinsic Scene Properties into Diffusion Models}\\}
  \vspace{1.0em}
  {\Large{Supplementary Material}}
  \vspace{1.0em}
\end{center}

\section{Theoretical Motivation of Utilizing Intrinsics to Reflect the Underlying Scene}
\label{sec:method-formulation}
\begin{figure}[h]
  \centering
  \includegraphics[width=0.5\columnwidth]{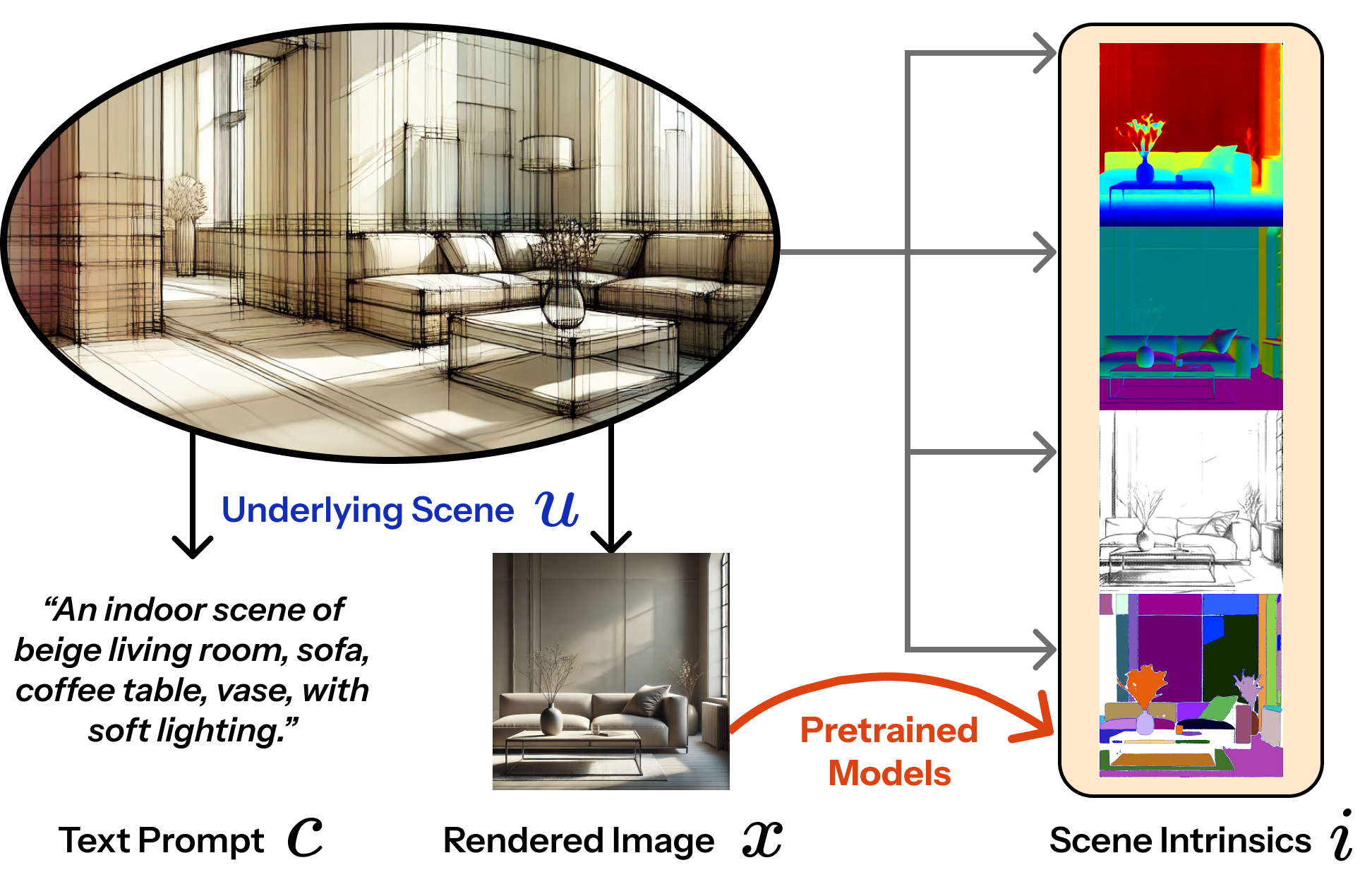}
  \caption{
  We assume that an image $x$ is constructed from a latent underlying scene $u$, 
  while text $c$ and intrinsic scene properties $i$ are different instantiations entailed by $u$, i.e., $u \to (x, i, c)$. 
  Instead of learning $u$ directly, which is difficult to obtain, we leverage pre-trained models to estimate various intrinsics that reflect $u$.
  }
  \label{fig:pgm}
\end{figure}

\subsection{Log-likelihood with Underlying Scene}

Given a dataset $(x, c) \sim D$ of image $x$ and text $c$ pairs, a typical generative model, such as DDPMs~\cite{ddpm}, aims to maximize the empirical log-likelihood:
\begin{align}
    \argmax_\theta \mathbb{E}_{(x, c) \sim D}[\log p_{\theta}(x|c)]. 
\end{align}
However, considering the latent underlying scene $u$ from which the data is built, 
we can consider a relationship as in \cref{fig:pgm}, where $x$ and $c$ are generated from the underlying scene $u$.
For low $p_\theta(x|u)$, the generated $x$ will not accurately reflect $u$, resulting in a spatially inconsistent image.
Thus, the objective of generating $x$ from $c$ with reflecting $u$ can be rewritten as:
\begin{align} \label{eq:new_obj}
\argmax_\theta\mathbb{E}_{(x, c) \sim D}\mathbb{E}_{p_\theta(u|c)}\left[\log p_{\theta}(x|u)\right], 
\end{align}
which is equivalent to the following (see \cref{appendix:math_equiv}):
\begin{align} \label{eq:new_obj_equiv}
\argmax_\theta\mathbb{E}_{(x, c) \sim D} &[\log p_{\theta}(x|c) - KL(p_\theta (u|c) || p_\theta (u|x, c))].
\end{align}
In other words, considering the underlying scene $u$, we aim to maximize the conventional log-likelihood while minimizing the Kullback-Leibler (KL) divergence between $p_\theta (u|c)$ and $p_\theta (u|x, c)$.
However, information on $u$ in $c$ is typically very limited and the KL term is intractable since $u$ is not directly observable. For this, we consider intrinsic scene properties that provide rich information about $u$.

\subsection{Introducing Intrinsic Scene Properties} \label{subsubsec:new_obj_intr}
In this work, we consider four intrinsic scene properties; depth map, surface normal, segmentation map, and line drawing, to reflect $u$ from various perspectives.
Notably, such intrinsics $i$ could be easily obtained using existing pre-trained estimators~\cite{metric3d, sam}, without requiring an additional layer of training or data collection.
With an augmented data set $(x, i, c) \sim D^+$, the objective can be written as follows:
\begin{align}  \label{eq:obj_intr}
\argmax_\theta
\mathbb{E}_{(x, i, c)}&_{\sim D^+} [\log p_{\theta}(x|i, c) - KL(p_\theta (u|i, c) || p_\theta (u|x, i, c))].
\end{align}
Our key insight is that the more information $i$ contains about $u$, the better the trained model can minimize KL divergence {(see \cref{appendix:math_ineq} for formal derivation)}.
Maximization of $p_{\theta}(x|i, c)$ implies optimizing the process of providing intrinsics as control elements to the image, such as ControlNet~\cite{controlnet}.
In practice, since our goal is not to provide $i$ to the text-to-image generation process, we instead maximize $\log p_\theta (x, i|c) = \log p_\theta(x|i, c) + \log p_\theta(i| c)$.

\section{Formal Derivation for \cref{eq:new_obj_equiv,eq:obj_intr}}
\label{appendix:math}

We first start by constructing a graphical model of $u, x, i$, and $c$.
Next, we show that \cref{eq:new_obj} is equivalent to \cref{eq:new_obj_equiv}.
Finally, we show that the more information $i$ contains about $u$, the lower the objective \cref{eq:obj_intr}.

\subsection{Graphical model with the underlying scene}
As illustrated in \cref{fig:pgm}, we consider an image $x$, text $c$, and intrinsic scene properties $i$ as different instantiations entailed by a latent underlying scene $u$. 
In other words, we can construct a probabilistic graphical model (PGM) relation $u \rightarrow x, i, c$.
Using Bayes' Rule, the following equation holds:
\begin{equation} \label{appdix_eq:pgm}
    p_{\theta}(u|x,c) = \frac{p_{\theta}(u,x,c)}{p_{\theta}(x,c)} = \frac{p_{\theta}(x|u)p_{\theta}(c|u)p_{\theta}(u)}{p_{\theta}(x,c)} 
    = \frac{p_{\theta}(x|u)p_{\theta}(u|c)p_{\theta}(c)}{p_{\theta}(x,c)} = \frac{p_{\theta}(x|u)p_{\theta}(u|c)}{p_{\theta}(x|c)}.
\end{equation}

\subsection{Equivalence between \cref{eq:new_obj} and \cref{eq:new_obj_equiv}} \label{appendix:math_equiv}
Starting from the KL divergence between $p_{\theta}(u|c)$ and $p_{\theta}(u|x,c)$ in \cref{eq:new_obj_equiv}, 
\begin{equation} \label{appdix_eq:KL}
    KL(p_{\theta}(u|c)||p_{\theta}(u|x,c)) = \int p_{\theta}(u|c)\log\frac{p_{\theta}(u|c)}{p_{\theta}(u|x,c)}du = \int p_{\theta}(u|c)\left[\log p_{\theta}(u|c)-\log p_{\theta}(u|x,c)\right] du.
\end{equation}
Substituting \cref{appdix_eq:pgm} into \cref{appdix_eq:KL},
\begin{align}
\begin{split}
    KL(p_{\theta}(u|c)||p_{\theta}(u|x,c)) &= \int p_{\theta}(u|c)\left[\log p_{\theta}(u|c)-(\log p_{\theta}(x|u)+\log p_{\theta}(u|c)-\log p_{\theta}(x|c))\right] du \\
    &= \int p_{\theta}(u|c)\left[-\log p_{\theta}(x|u)+\log p_{\theta}(x|c)\right] du \\
    &= -\mathbb{E}_{p_{\theta}(u|c)}\left[\log p_{\theta}(x|u)\right]+\log p_{\theta}(x|c).
\end{split}
\end{align}
Thus,
\begin{equation} \label{appdix_eq:final}
    \mathbb{E}_{p_\theta(u|c)}\left[\log p_{\theta}(x|u)\right] = 
    \log p_{\theta}(x|c) - KL(p_\theta (u|c) || p_\theta (u|x, c)).
\end{equation}
which indicates the equivalence between \cref{eq:new_obj} and \cref{eq:new_obj_equiv}.
By replacing $c$ with $(i, c)$, we can get $\mathbb{E}_{p_\theta(u|i, c)}\left[\log p_{\theta}(x|u)\right] = \log p_{\theta}(x|i, c) - KL(p_\theta (u|i, c) || p_\theta (u|x, i, c))$ in the same way.

\subsection{Inequality between \cref{eq:new_obj_equiv} and \cref{eq:obj_intr}} \label{appendix:math_ineq}
To support our claim in Line 210, we first show that the inequality
\begin{align}
    KL(p_\theta(u|c)||p_\theta(u|x,c))\geq \mathbb{E}_{p_\theta (i|c)} KL(p_\theta(u|i,c)||p_\theta(u|x,i,c)
\end{align}
holds. This can be proved by the chain rule for KL divergence and the conditional independence $i \perp (x,c) | u$.

Applying the KL chain rule 
\begin{align}
    KL(p(x,y)||q(x,y))=KL(p(y)||q(y))+\mathbb{E}_{p(y)}[KL(p(x|y)||q(x|y))]
\end{align}
in $KL(p_\theta(u,i|c)||p_\theta(u,i|x,c)$ with decomposing by conditioning on $i$ and on $u$ respectively,
\begin{align}
    KL(p_\theta(u,i|c)||p_\theta(u,i|x,c)) &= \mathbb{E}_{p_\theta (i|c)}[KL(p_\theta(u|i,c)||p_\theta(u|x,i,c))] + KL(p_\theta(i|c)||p_\theta(i|x,c)) \label{appdix_eq:chain1} \\
    KL(p_\theta(u,i|c)||p_\theta(u,i|x,c)) &= \mathbb{E}_{p_\theta (u|c)} [KL(p_\theta(i|u,c)||p_\theta(i|x,u,c))] + KL(p_\theta(u|c)||p_\theta(u|x,c)). \label{appdix_eq:chain2}
\end{align}
Then, applying the conditional independence $i \perp (x,c) | u$ implied by the relation $u\rightarrow x,i,c$,
\begin{align}
    p_\theta(i|u,x,c)&=p_\theta(i|u) \\
    p_\theta(i|u,c)&=p_\theta(i|u),
\end{align}
$KL(p_\theta(i|u,c)||p_\theta(i|x,u,c))=KL(p_\theta(i|u)||p_\theta(i|u))=0$ in \cref{appdix_eq:chain2}. Since $KL(p_\theta(i|c)||p_\theta(i|x,c))\geq 0$ in \cref{appdix_eq:chain1},
\begin{align}
\begin{split}
    KL(p_\theta(u,i|c)||p_\theta(u,i|x,c)) 
    &= \mathbb{E}_{p_\theta (u|c)} [\cancel{KL(p_\theta(i|u,c)||p_\theta(i|x,u,c))}] + KL(p_\theta(u|c)||p_\theta(u|x,c)) \\
    &= 0+KL(p_\theta(u|c)||p_\theta(u|x,c)) \\
    &= \mathbb{E}_{p_\theta (i|c)}[KL(p_\theta(u|i,c)||p_\theta(u|x,i,c))] + KL(p_\theta(i|c)||p_\theta(i|x,c)) \\
    &\geq \mathbb{E}_{p_\theta (i|c)}[KL(p_\theta(u|i,c)||p_\theta(u|x,i,c))].
\end{split}
\end{align}
Hence,
\begin{align}
    KL(p_\theta(u|c)||p_\theta(u|x,c))\geq \mathbb{E}_{p_\theta (i|c)}[KL(p_\theta(u|i,c)||p_\theta(u|x,i,c))].
\end{align}

Applying for $u\rightarrow i_1, i_2, ... i_k$ in the same way, we can get:
\begin{align}
\begin{split}
    KL(p_\theta(u|c)||p_\theta(u|x,c))
    &\geq \mathbb{E}_{p_\theta (i_1|c)}[KL(p_\theta(u|i_1,c)||p_\theta(u|x,i_1,c))] \\
    &\geq \mathbb{E}_{p_\theta (i_1,...,i_k|c)}[KL(p_\theta(u|i_1,...,i_k,c)||p_\theta(u|x,i_1,...,i_k,c))],
\end{split}
\end{align}
which supports our claim that with more information $i=i_1,..,i_k$, the lower the objective \cref{eq:obj_intr}.

\section{Experimental Details}
\label{appendix:exp-details}

\subsection{Dataset}
For the image dataset, we use a subset of 542k images with aesthetic scores of 6.5 or higher from LAION-5B~\cite{laion}.
We use Metric3D~\cite{metric3d} with a ViT-L backbone to estimate the depth map and surface normal.
Following the method of Marigold~\cite{marigold}, we normalized each depth map by mapping its 2\% and 98\% percentiles to -1 and 1, respectively.
We use an automatic mask generator with \textit{sam2.1\_hiera\_base\_plus} model~\cite{sam} to estimate segmentation maps, 
with \textit{pred\_iou\_thresh} 0.7, \textit{stability\_score\_thresh} 0.92, \textit{stability\_score\_offset} 0.7, \textit{crop\_n\_layers} 1, and \textit{crop\_n\_points\_downscale\_factor} 2.
To estimate line drawings, we use a publicly available image-to-line drawing model\footnote{\url{https://huggingface.co/spaces/awacke1/Image-to-Line-Drawings/blob/main/model.pth}}.
All intrinsics are normalized to a range of $[-1, 1]$.

\subsection{Training Intrinsic VAE}
For the input and output of the Intrinsic VAE, all intrinsics are encoded in RGB format (total $4\times 3$ channels).
We train intrinsic VAE with a batch size of 256 for 2000 steps with a learning rate of $1e-4$ using the AdamW optimizer for each base model, with resolution $768\times 768$ for SD2.1 and $1024\times 1024$ for SDXL.
After 200 training steps, we apply LPIPS loss on a scale of 1.0 and discriminator loss on a scale of 0.5.
The discriminator uses a 5-layer CNN architecture. 
To improve training stability, random noise is added to the fake images during the initial training stages, and the noise proportion is gradually reduced to zero over the first 160 steps.
A colormap is applied to the grayscale depth maps to ensure that different depth levels are perceptually distinguishable.

\subsection{Training \model{}}
To train the denoising process in the intrinsic domain, we adapt LoRA for the self-attention layers with rank 32, training with a batch size of 128 for 5000 steps with a learning rate of $2e$-$4$ using the AdamW optimizer.
We set $\lambda=4$ in \cref{eq:loss}.
Since the surface normal estimator occasionally provides a flat surface for images such as paintings, we discarded the data with a surface normal standard deviation of 0.03 or less when training the UNet. 
The resolution of the images for training is fixed at 512 using the same random cropping method suggested in \citet{sdxl}, while we sample the image with a resolution of $768\times 768$.
For the loss function \cref{eq:loss} with SD2.1, we use the corresponding loss function with v-prediction~\cite{v_prediction} instead of $\epsilon$-prediction.
We selected hyperparameters based on prior works and rule of thumb, rather than extensive tuning, due to the high cost of each experimental run.

For weight scheduling, we assign layer indices $l$ as follows: 
SD2.1 consists of four down-blocks, four corresponding up-blocks, and one mid-block.
We assign indices 1 through 4 from the early block to the four down-blocks, 5 to the mid-block, and 6 through 9 to the up-blocks.
We denote timestep $t$ as the range of DDPM [0, 1000]~\cite{ddpm}.
For the hyperparameters in weight scheduling mechanism (\cref{eq:schedule}), we tested $\tau$ and $\sigma$ in increments of 100, and tested $\alpha$ with values of 0.5 and 1.0. 

\subsection{Metrics}
The CMMD values~\cite{cmmd} reported in \cref{table:general,table:robustness,table:ablation} are computed by comparing generated images with images from the base model.
The CMMD values in \cref{table:hand} are computed by comparing the generated images with the 3,493 hand images from the InterHand dataset used during training.
For HPSv2~\cite{hpsv2}, we use the HPS v2.1 model for evaluation.
To evaluate the AI preference for spatial inconsistency using GPT-4o, we use the following procedure. 
First, we provide the model with two images generated by the base model and \model{} with the same prompt and seed, horizontally aligned with a separator line and vertically resized to 512 resolution.
We then provide the instruction in the following format, similar to the prompts proposed by PixArt-$\Sigma$~\cite{pixart_sigma}:
\textit{"As an AI visual assistant, you are analyzing two images arranged to the left and right of a given image. When presented with a specific caption, it is required to evaluate and determine whether the two images are equally good or which image better reflects the actual physical scene. Please pay attention to the key information, probable architectures, consistency for occlusion, avoidance of paradoxical visual effects, correct object shapes and structures, and anatomical accuracies. The caption for the two images is: \{prompt\}".
"Please respond me strictly in the following format: \textless The left image is better\textgreater~or \textless The right image is better\textgreater~or \textless The two images are tied\textgreater. The reason is (give your reason here)."}

\subsection{Network Configuration.}
\begin{table}[t]
    \centering
    \caption{Model configuration (A6000 GPU, fp32, single image).}
    \begin{adjustbox}{width=0.4\linewidth}
    \begin{tabular}{lccc}
        \toprule
        & \# params & inference time & memory \\ \midrule
        \multicolumn{1}{c}{\multirow{2}{*}{Base}} & 865M (UNet)    & 8.76s      & \multirow{2}{*}{6.65 GiB} \\
        \multicolumn{1}{c}{}                      & 83.6M (VAE)     & 17.2s ($\times$2) &   \\
        \midrule
        \multirow{2}{*}{Ours}                     & 872M (UNet)    & 28.9s (Drop)    & \multirow{2}{*}{7.37 GiB} \\
        & 167M (VAE)     & 19.0s (Gauss)   &                   \\ \bottomrule
    \end{tabular}
    \end{adjustbox}
    \label{table:config}
\end{table}

\cref{table:config} summarizes the parameter sizes, time, and memory requirements for generating single images.
Note that for Gaussian weight scheduling, we disable the denoising process for steps in $t\leq 500$, since it only uses information from the intrinsic domain near $\tau=800$.
Since \model{} generates images and intrinsics simultaneously, the time consumption is higher than for the base model.
However, the comparison between Base $\times$2 and \model{} (Gauss) in \cref{table:general,table:hand} shows that our model produces more consistent images with similar time consumption.
It also reveals that the number of additional parameters and memory size is about 10\% of the base model.

\subsection{Implementation details}


To efficiently co-generate images and intrinsics, we leverage our architecture design in which the intrinsic LDM corresponds to the image LDM augmented only with LoRA parameters.
By selectively enabling or disabling these LoRA components, we switch between the image and intrinsic domains within a single shared model.
For implementation, we first concatenate the image and intrinsic latents batch-wise.
Inside each attention processor of the LDM, we split the batch into two halves to compute queries, keys, and values of each domain. 
For the first half of the batch, LoRA weights are not applied ($Q_x\cdot z_x, K_x\cdot z_x, V_x\cdot z_x$), while for the second half, LoRA weights are applied ($Q_i\cdot z_i, K_i\cdot z_i, V_i\cdot z_i$). 
Keys and values of both domains are then concatenated feature-wise to form cross-domain co-generation as in \cref{eq:cross-domain-self-attention}.
Finally, the queries, keys, and values from both domains are concatenated batch-wise and passed to the attention operation.

During inference, classifier-free guidance (CFG) is applied to both the image and intrinsic domains, requiring four batch elements per sample. 
The weight scheduling mechanism introduced in \cref{sec:method-ildm} is used, passing the current layer index $l$ and denoising timestep $t$ to each attention processor.
The processor computes the weight matrix $W^{(l,t)}$ from \ref{eq:attention}, which is added as a weight bias in the attention computation.
The inference pipeline is summarized in \cref{appdix_alg:inference_pipeline}.

\begin{algorithm}[t]
\label{appdix_alg:inference_pipeline}
\caption{Inference Pipeline of I-LDM}
\begin{algorithmic}[1]
\REQUIRE scheduler timesteps $\mathcal{T}$, CFG scale $s$, prompt and negative prompt embeddings $p, n$.
\STATE Sample noise $\epsilon \sim \mathcal{N}(0,I)$.
\STATE Initialize latent batch $z = [\epsilon,\epsilon,\epsilon,\epsilon]$ and condition batch $c=[n,p,n,p]$.
\FOR{each $t \in \mathcal{T}$}
    \STATE Set input to model: $(z, t, c)$.
    \FOR{each transformer layer $l=1,\dots,L$}
        \STATE \textbf{Attention processor:} 
        \STATE \quad \textbf{Receives} $z^{(l)} = \left[z_x^{(l,n)},\, z_x^{(l,p)},\, z_i^{(l,n)},\, z_i^{(l,p)}\right]$, weights $Q_x^{(l)}, K_x^{(l)}, V_x^{(l)}$, LoRA weights $Q'^{(l)}, K'^{(l)}, V'^{(l)}$.
        \STATE \quad Compute $Q^{(l)}_x\cdot z_x^{(l,\cdot)}, K^{(l)}_x\cdot z_x^{(l,\cdot)}, V^{(l)}_x\cdot z_x^{(l,\cdot)}$ with \textit{disabling} LoRA.
        \STATE \quad Compute $\left(Q^{(l)}_x+Q'^{(l)}\right)\cdot z_i^{(l,\cdot)}, \left(K^{(l)}_x+K'^{(l)}\right)\cdot z_i^{(l,\cdot)}, \left(V^{(l)}_x+V'^{(l)}\right)\cdot z_i^{(l,\cdot)}$ with \textit{enabling} LoRA.
        \STATE \quad Compute $q_\cdot, k_\cdot, v_\cdot$ (\cref{eq:cross-domain-self-attention}) and $W^{(l,t)}$ (\cref{sec:method-ildm})
        \STATE \textbf{Compute attention:} $Attn_x^{(l, t)}, Attn_i$ (\cref{eq:attention})
    \ENDFOR
    \STATE Obtain model outputs $\left[\epsilon^{(n)}_x,\ \epsilon^{(p)}_x,\ \epsilon^{(n)}_i,\ \epsilon^{(p)}_i\right]$.
    \STATE \textbf{Classifier-free guidance (per-domain):}
    \STATE \quad $\hat{\epsilon}_x \leftarrow \epsilon^{(n)}_x + s\big(\epsilon^{(p)}_x - \epsilon^{(n)}_x\big)$
    \STATE \quad $\hat{\epsilon}_i \leftarrow \epsilon^{(n)}_i + s\big(\epsilon^{(p)}_i - \epsilon^{(n)}_i\big)$
    \STATE $z \leftarrow \text{SchedulerUpdate}\left(z, [\hat{\epsilon}_x,\,\hat{\epsilon}_x,\,\hat{\epsilon}_i,\,\hat{\epsilon}_i], t\right)$
\ENDFOR
\RETURN $z_x$.
\end{algorithmic}
\end{algorithm}

\subsection{Computing Infrastructure}
Experiments are conducted on a workstation equipped with 2× NVIDIA RTX A6000 GPUs, AMD Ryzen Threadripper 3960X 24-Core CPU, and 256GB of RAM, running Ubuntu 22.04.4 LTS. We use PyTorch 2.4.1 with CUDA 12.3 as the primary deep learning framework, and employ diffusers v0.28.0 and transformers v4.44.2 for implementation.

\section{Latent Distribution for Images and Intrinsics} \label{appendix:latent}

\begin{figure}[!h]
    \centering
    \subfloat[Visualizations of latent distributions with t-SNE.]{
        \begin{adjustbox}{valign=b, width=0.4\linewidth}
            \includegraphics{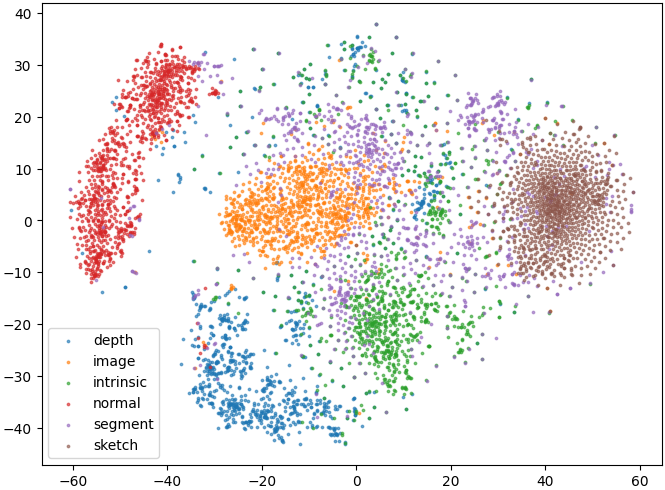}
        \end{adjustbox}
        \label{appdix_fig:tsne}
    }
    \quad
    \subfloat[MMD between image and intrinsic.]{
        \begin{adjustbox}{valign=b, width=0.28\linewidth}
            \begin{tabular}{lr}
                \toprule
                \textbf{Intrinsic} & \textbf{MMD} \\ \midrule
                Intrinsic VAE   & \textbf{7.921}        \\ 
                Depth map       & 8.225        \\ 
                Surface normal  & 24.349        \\ 
                Line drawings   & 44.401        \\ 
                Segmentation map& 8.893         \\ \bottomrule
            \end{tabular}
        \end{adjustbox}
        \label{appdix_fig:mmd}
    }
    \caption{Quantitative and qualitative analysis of the latent distribution of each intrinsic and intrinsic VAE.}
\end{figure}

As described in \cref{sec:method-ise}, if the latent distribution of the intrinsic features encoded by the VAE differs significantly from that of the images, challenges may arise during the training and sampling processes. 
\cref{appdix_fig:tsne} visualizes these distributions using t-SNE: it shows the distribution of 1,000 images from the LAION dataset encoded by the pre-trained VAE (orange); the distributions of each intrinsic encoded by the pre-trained VAE, depth maps (blue), surface normals (red), line drawings (purple), and segmentation maps (brown); and the distribution of all intrinsics encoded by our intrinsic VAE (green).
\cref{appdix_fig:mmd} presents a table listing the Maximum Mean Discrepancy (MMD) values, calculated using a Gaussian kernel, between the latent distributions of the images and each intrinsic. 
The results show that certain intrinsics, such as surface normals and line drawings, are far from the image distribution. 
In contrast, the distribution of the intrinsics encoded by the intrinsic VAE closely resembles that of the images.

\section{Adapting \model{} into SDXL and PixArt-$\alpha$}
\subsection{Training details}
For SDXL, we adapt LoRA weights to self and cross-attention layers with rank 8, with batch size 64 for 10000 steps with a learning rate of $2e-5$, while other settings are equal to SD2.1. 
The resolution of the images for training is fixed at 512, while we sample the image with a resolution of $1024\times 1024$.
For sampling, the cfg scale is set to 7.5.
For weight scheduling, we assign layer indices $l$ as follows: 
SDXL consists of three down-blocks, three corresponding up-blocks, and one mid-block; we assign indices 1 through 3 from the early block to the three down-blocks, 4 to the mid-block, and 5 through 7 to the up-blocks.
For generating samples from Multi prompts, we set $L=\{4\}, \tau=900$ for $w_{d}$, $\alpha=0.5, \tau=800, \sigma=50$ for $w_{g}$.
For complex hand structure generation task, we set $L=\{3,4,5\}, \tau=900$ for $w_d$, $\alpha=1, \tau=800, \sigma=100$ for $w_g$.

For PixArt-$\alpha$, we use PixArt-$\alpha$-512 model as the base model.
We adapt LoRA weights to self-attention layers with rank 32, with batch size 128 for 2000 steps with a learning rate of $2e-6$, while other settings are equal to SD2.1. 
The resolution of the images for training is fixed at 384, while we sample the image with a resolution of $512\times 512$.
For sampling, the cfg scale is set to 4.5.
PixArt-$\alpha$ consists of 28 Transformer blocks, indicated by 1 through 28 from the first block. 
For generating samples from Multi prompts, we set $L=[1, 28], \tau=900$ for $w_{d}$, $\alpha=1, \tau=800, \sigma=0.3$ for $w_{g}$.

\subsection{Quantitative Results}
\label{appendix:quantitative}
\begin{table*}[h]
    \centering
    \caption{
    Quantitative results on generated images in Multi prompts based on SDXL and PixArt-$\alpha$. 
    The evaluation is done with images sampled from 3 seeds for each prompt. 
    The best result is bolded. 
    }
    \begin{adjustbox}{width=\textwidth}
    \begin{tabular}{lccggbbccggbb}
    \toprule
                & \multicolumn{6}{c}{\begin{tabular}[c]{@{}l@{}l@{}l@{}l@{}l@{}}SDXL \citep{sdxl}\end{tabular}} & \multicolumn{6}{c}{\begin{tabular}[c]{@{}l@{}l@{}l@{}l@{}l@{}}PixArt-$\alpha$ \citep{pixart}\end{tabular}} \\
    \cmidrule(lr){2-7} \cmidrule(lr){8-13}
     & & & \multicolumn{2}{c}{\begin{tabular}[c]{@{}l@{}}human preference estimator\end{tabular}} & \multicolumn{2}{c}{\begin{tabular}[c]{@{}l@{}}LLM (GPT-4o)\end{tabular}} & & & \multicolumn{2}{c}{\begin{tabular}[c]{@{}l@{}}human preference estimator\end{tabular}} & \multicolumn{2}{c}{\begin{tabular}[c]{@{}l@{}}LLM (GPT-4o)\end{tabular}} \\
     \cmidrule(lr){4-5} \cmidrule(lr){6-7} \cmidrule(lr){10-11} \cmidrule(lr){12-13}
    Method          & CMMD($\downarrow$)      & CLIP($\uparrow$)      & ImageReward($\uparrow$)   & HPSv2($\uparrow$)     & win($\uparrow$) & lose($\downarrow$)
                    & CMMD($\downarrow$)      & CLIP($\uparrow$)      & ImageReward($\uparrow$)   & HPSv2($\uparrow$)     & win($\uparrow$) & lose($\downarrow$) \\ \midrule
    Base            & -     & 0.2876    & -0.1136       & 0.2624    & - & -
                    & -     & 0.2707    & -0.0665       & 0.2891    & - & -   \\
    \textbf{\model{}} (Drop)      & 0.01729     & 0.2892    & -0.0484      & 0.2624    & 31.8\% & 15.7\%
                    & 0.0044     & \textbf{0.2718}    & \textbf{-0.0428}      & \textbf{0.2899}    & \textbf{21.3\%} & \textbf{12.2\%}   \\
    \textbf{\model{}} (Gauss)     & 0.02694     & \textbf{0.2909}    & \textbf{-0.0469}      & \textbf{0.2634}    & \textbf{40.6\%} & \textbf{14.1\%}
                    & 0.0032     & 0.2713    & -0.0522      & 0.2897    & 19.0\% & 13.0\% \\
\bottomrule
\end{tabular}
\end{adjustbox}
\label{appdix_table:general}
\end{table*}

\begin{table}[h]
    \centering
    \caption{
    Quantitative results on 1000 generated images based on SDXL and PixArt-$\alpha$ trained with InterHand samples. 
    The best result is bolded.
    }
    \begin{adjustbox}{width=0.9\textwidth}
    \begin{tabular}{lccggccgg}
    \toprule
                & \multicolumn{4}{c}{\begin{tabular}[c]{@{}l@{}l@{}l@{}}SDXL \citep{sdxl}\end{tabular}} & \multicolumn{4}{c}{\begin{tabular}[c]{@{}l@{}l@{}l@{}}PixArt-$\alpha$ \citep{pixart}\end{tabular}} \\
    \cmidrule(lr){2-5} \cmidrule(lr){6-9}
    Method          & CMMD($\downarrow$)      & HPSv2($\uparrow$)     & Detection($\uparrow$) & Confidence($\uparrow$) &
    CMMD($\downarrow$) & HPSv2($\uparrow$)     & Detection($\uparrow$) & Confidence($\uparrow$)
    \\ \midrule  
    Base             & 0.9140    & \textbf{0.1640}    & 57.03\%   & 0.9181 & \textbf{1.2529}     & 0.1734    & 77.00\%   & 0.8704    \\
    \textbf{\model{}} (Drop)     & \textbf{0.7690}    & 0.1636    & 72.40\%   & 0.9301   & 1.2774     & \textbf{0.1785}    & 89.00\%   & \textbf{0.8721} \\
    \textbf{\model{}} (Gauss)    & 0.8554    & 0.1621    & \textbf{74.10\%}   & \textbf{0.9372}  & 1.2707     & 0.1732    & \textbf{92.20\%}   & 0.8688  \\
\bottomrule
\end{tabular}
\end{adjustbox}
\label{appdix_table:hand}
\end{table}

\cref{appdix_table:general,appdix_table:hand} summarizes the quantitative results for Multi prompts and InterHand experiment based on SDXL and PixArt-$\alpha$. The qualitative results, including the generated intrinsic, are visualized in \cref{appendix:qualitative}.
This shows that our method is also applicable to more scalable base models and other architectures.
However, the decrease in improvement over SD2.1 for the human preference estimator may be due to the fact that as the scalability of the base model increases, the score becomes more dominated by other evaluation factors (e.g., human aesthetics, text alignment) as the spatial inconsistency becomes less pronounced.
On the other hand, the win-loss ratio of GPT-4o and the high detection rate and confidence of the InterHand experiment, which excludes other evaluation factors besides spatial inconsistency, show that our model produces more spatially consistent images.

\section{Additional Analysis for the Robustness to Intrinsic Errors} \label{appendix:robustness}

In this section, we perform further quantitative and qualitative analysis of the relation between intrinsic and image quality.

\begin{figure}[h]
  \centering
  \includegraphics[width=0.73\linewidth]{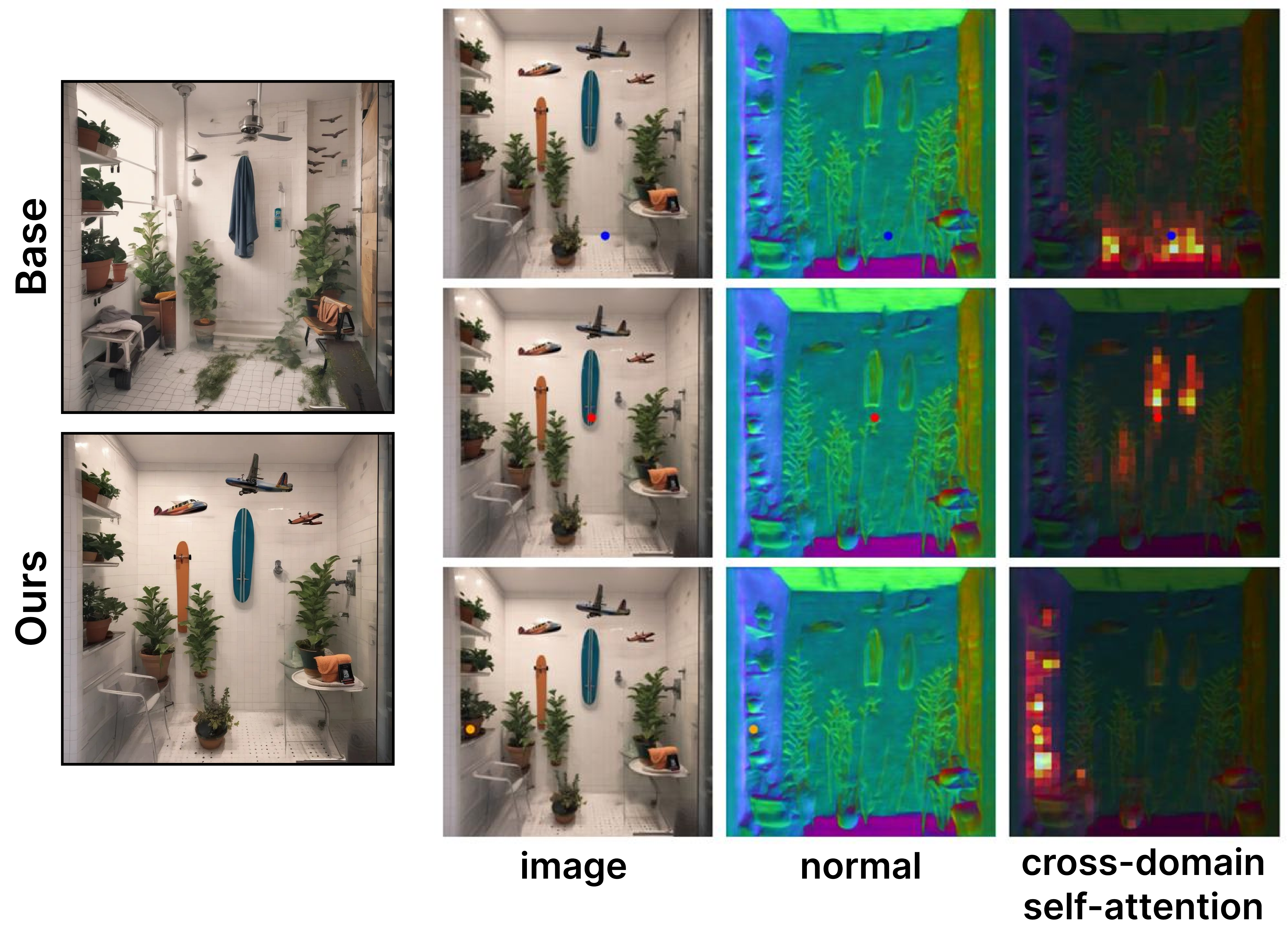}
  \caption{
  Visualizations of the generated image (Left) and surface normal (Middle) from \model{} (Drop) based on SDXL with prompt "potted plants and chairs and airplanes and snowboards in a shower.". (Right) The heatmap of cross-domain self-attention in colored dot areas where the image domain references from the intrinsic domain.
  Cross-domain self-attention to corners of walls and floors (Top), snowboards (Middle), and pots (Bottom).
  }
  \label{appdix_fig:attention}
\end{figure}

\begin{table}[h]
    \centering
    \caption{
    RMS error for generated depth map and mean $^\circ$ error for surface normal for Multi prompts. 
    }
    \begin{adjustbox}{width=0.6\linewidth}
        \begin{tabular}{llcccc}
        \toprule
        Base       & Schedule      & Depth RMSE ($\downarrow$)   & Mean $^\circ$ ($\downarrow$)    & ImageReward($\uparrow$)   & HPSv2($\uparrow$) \\ \midrule 
        \multirow{3}{*}{SD2.1} 
                    & Drop      & 0.2974    & 14.66     & -0.6148       & 0.2461    \\
                    & Gaussian  & 0.3059    & 15.48     & -0.5739       & 0.2450    \\
                    & X         & 0.2773    & 13.78     & -0.5191       & 0.2427    \\
        \midrule
        \multirow{3}{*}{SDXL} 
                    & Drop      & 0.3664    & 18.52     & -0.0484       & 0.2624    \\
                    & Gaussian  & 0.3486    & 18.40     & -0.0469       & 0.2634    \\
                    & X         & 0.2455    & 15.34     & -0.4483       & 0.2432    \\
        \bottomrule
        \end{tabular}
    \end{adjustbox}
    \label{appdix_table:mse}
\end{table}

To evaluate the accuracy of the generated intrinsics, we estimate depth maps and surface normals from the generated images using Metric3D and compare the errors between the generated intrinsics.
\cref{appdix_table:mse} shows the intrinsic accuracy and human preference estimatior results of \model{} with Drop, Gaussian, and no weight scheduling.
Here, depth RMSE denotes the root mean square error of the depth maps, and Mean $^\circ$ denotes the average angular error in degrees between the surface normals. 
This is comparable in accuracy to the existing work ~\cite{intrinsic_lora}, which trained LoRA to generate intrinsic from Stable Diffusion (while the difference in scale in depth RMSE is due to differences in depth encoding).
Removing weight scheduling can slightly reduce the misalignment between image and intrinsic, but this results in a degradation of image quality and the generation of unnatural object layouts, as visualized in ~\cref{fig:schedule}.

We further verify that the cross-domain self-attention mechanism can convey meaningful information even in the presence of cross-domain misalignment.
\cref{appdix_fig:attention} illustrates the regions in the intrinsic domain that the image domain references through cross-domain self-attention in when the alignment between domains is not perfectly matched.
In the example shown in the first row, the blue dot is located at the corner between the wall and the floor in the image. 
In the generated surface normal map, however, the same position points to the wall instead of the corner. 
Despite this misalignment, the cross-domain self-attention heatmap is concentrated around the corner region in the surface normal map.
Similarly, in the examples shown in the second and third rows, the colored dots are placed on the snowboard and the pot, respectively, in the image, but do not correspond to the same objects in the intrinsic domain. 
Nevertheless, the heatmaps show that attention is focused on semantically similar regions.
These observations support our claim that meaningful information can be exchanged by focusing on semantically similar regions, even when spatial misalignments exist.

\begin{table}[t]
    \centering
    \subfloat[Differences by challenge.]{
        \begin{adjustbox}{width=0.49\linewidth}
        \begin{tabular}{lcccccc}
            \toprule
                    & \multicolumn{3}{c}{\model{} (Gaussian)}     & \multicolumn{3}{c}{(\model{} (Drop)} \\
                    \cmidrule(lr){2-4} \cmidrule(lr){5-7}
    Challenge       & $\Delta$CLIP & $\Delta$ImageReward & $\Delta$HPSv2 & $\Delta$CLIP & $\Delta$ImageReward & $\Delta$HPSv2  \\
            \midrule
Basic               & \red{-0.0011} & 0.0657            & 0.0071    & \red{-0.0018} & \red{-0.0005}     & 0.0050 \\
Complex             & 0.0031        & 0.1566            & 0.0104    & \red{-0.0001} & 0.1112            & 0.0093 \\
Fine-grained Detail & 0.0023        & 0.1418            & 0.0089    & 0.0009        & 0.1061            & 0.0069 \\
Imagination         & 0.0007        & 0.2368            & 0.0096    & \red{-0.0052} & 0.1140            & 0.0059 \\
Linguistic Structures & \red{-0.0009} & 0.1290          & 0.0061    & \red{-0.0039} & 0.0502            & 0.0037 \\
Perspective         & 0.0039        & 0.2436            & 0.0114    & 0.0033        & 0.1588            & 0.0072 \\
Properties \& Positioning & 0.0068  & 0.2810            & 0.0073    & 0.0048        & 0.2685            & 0.0097 \\
Quantity            & \red{-0.0004} & 0.0092            & 0.0072    & \red{-0.0012} & \red{-0.0053}     & 0.0052 \\
Simple Detail       & 0.0010        & 0.0659            & 0.0068    & \red{-0.0005} & 0.0212            & 0.0043 \\
Style \& Format     & 0.0012        & 0.2600            & 0.0108    & \red{-0.0023} & 0.1811            & 0.0098 \\
Writing \& Symbols  & 0.0033        & 0.1137            & 0.0084    & 0.0008        & 0.1037            & 0.0045 \\
            \bottomrule
        \end{tabular}
        \end{adjustbox}
        \label{appdix_table:parti_challenge}
    }
    \subfloat[Differences by category.]{
        \begin{adjustbox}{width=0.49\linewidth}
        \begin{tabular}{lcccccc}
            \toprule
                    & \multicolumn{3}{c}{\model{} (Gaussian)}     & \multicolumn{3}{c}{(\model{} (Drop)} \\
                    \cmidrule(lr){2-4} \cmidrule(lr){5-7}
Category            & $\Delta$CLIP & $\Delta$ImageReward & $\Delta$HPSv2 & $\Delta$CLIP & $\Delta$ImageReward & $\Delta$HPSv2  \\
            \midrule
Abstract            & \red{-0.0051} & 0.0220            & 0.0087    & \red{-0.0032} & \red{-0.0410}     & 0.0097 \\
Vehicles            & 0.0018        & 0.1519            & 0.0073    & 0.0000        & 0.1017            & 0.0076 \\
Illustrations       & 0.0034        & 0.1499            & 0.0102    & \red{-0.0002} & 0.1037            & 0.0090 \\
Arts                & \red{-0.0008} & 0.1321            & 0.0029    & \red{-0.0080} & 0.0152            & 0.0064 \\
World Knowledge     & 0.0018        & 0.2430            & 0.0108    & \red{-0.0015} & 0.1706            & 0.0087 \\
People              & 0.0008        & 0.2226            & 0.0117    & \red{-0.0016} & 0.1230            & 0.0083 \\
Animals             & 0.0015        & 0.1724            & 0.0085    & \red{-0.0005} & 0.1164            & 0.0036 \\
Artifacts           & \red{-0.0006} & 0.0423            & 0.0059    & \red{-0.0009} & 0.0151            & 0.0037 \\
Food \& Beverage    & 0.0060        & 0.0612            & 0.0084    & 0.0011        & 0.0091            & 0.0051 \\
Produce \& Plants   & 0.0004        & 0.0065            & 0.0059    & \red{-0.0018} & \red{-0.0454}     & 0.0029 \\
Outdoor Scenes      & 0.0034        & 0.1730            & 0.0105    & 0.0017        & 0.1054            & 0.0098 \\
Indoor Scenes       & 0.0003        & 0.0160            & 0.0066    & 0.0003        & 0.0789            & 0.0065 \\
            \bottomrule
        \end{tabular}
        \end{adjustbox}
        \label{appdix_table:parti_category}
    }
    \caption{Difference in metrics between SD2.1 and \model{} for each (a) challenge and (b) category in Parti prompt.
    The red numbers indicate cases where the metric of \model{} is low, while others represent equal or high metric.}
\end{table}

\begin{figure}[!h]
  \centering
  \includegraphics[width=0.98\linewidth]{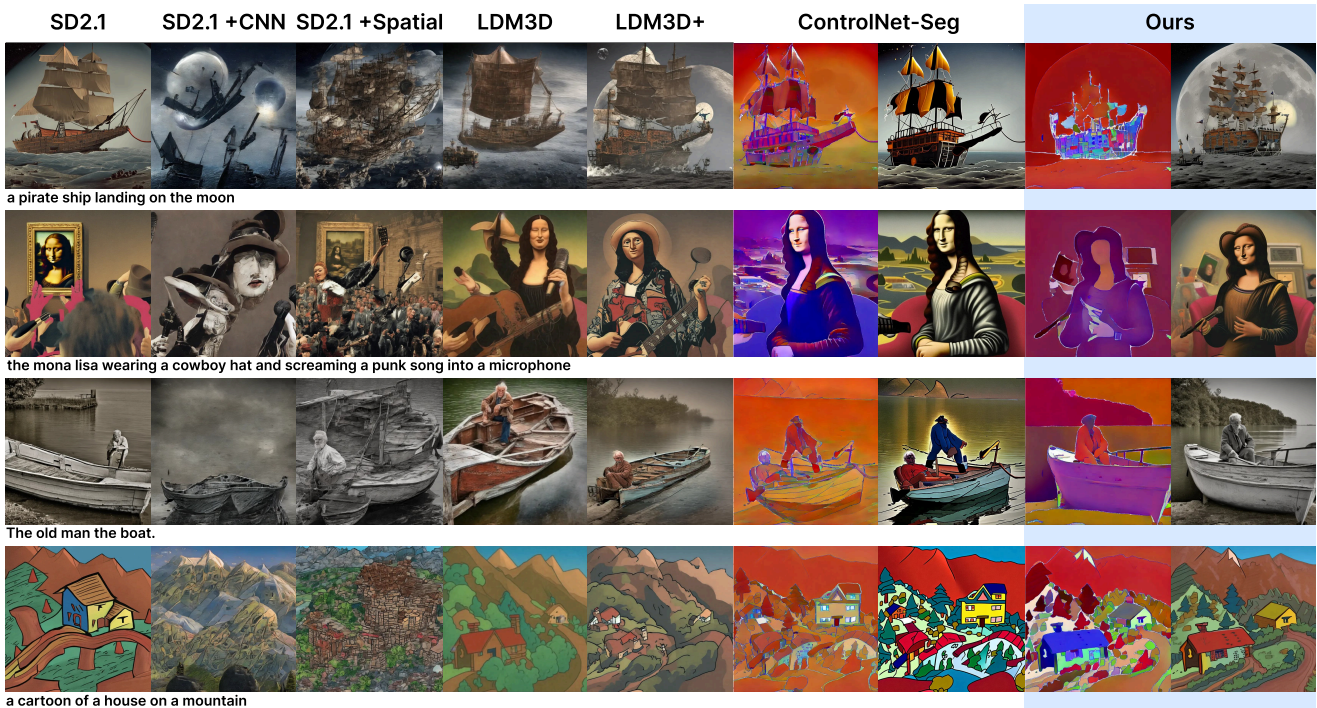}
  \caption{
  Visualizations of samples from baseline approaches, SD2.1, SD2.1 +CNN, SD2.1 +SpatialAdd, LDM3D, LDM3D+, ControlNet-Seg with generated segmentation map,
  and ours.
  }
  \label{appdix_fig:ablation_method}
\end{figure}

\section{Qualitative Results and Further Analysis in T2I Generation}

\subsection{Analysis of metrics by category and challenge for Parti prompts} \label{appendix:diversity}

\cref{appdix_table:parti_challenge} and \cref{appdix_table:parti_category} summarize the performance difference between the base model and ours for each challenge and category of Parti prompts. 
Overall, we observe that the scores of the human preference estimators are substantially higher in all categories and challenges.
In particular, scores are maintained or even improved in the Abstract and Art categories, which involve the representation of abstract scenes, and in the Style \& Format category, which includes various rendering styles. 
This suggests that our method preserves the base model's capability to manage general textual conditions.

\subsection{Comparisons between Baseline Methods}
\label{appendix:comparison_baseline}

\cref{appdix_fig:ablation_method} illustrates a qualitative comparison between our method and the baseline approaches introduced in the baseline section of \cref{sec:exp}. These methods have difficulty resolving spatial inconsistencies and significantly degrade the quality of the base model.
Our analysis of each method is as follows:

\textbf{1. Adding additional input, output channels to the UNet (+CNN, +SpatialAdd)}: 
The main reason for the failure of simply adding CNN channels is that, unlike the base model, our work also requires the generation of intrinsics.
Since changes to the up-block representations are necessary, the quality of the image domain degrades by simply fine-tuning the T2I model.
The same problem also occurs with +SpatialAdd.
Unlike ControlNet, which only uses intrinsics as inputs, generating intrinsics requires changes in Up-block representations.
Thus, fine-tuning the final convolution layer or the whole Up-block network is required, which inevitably degrades image quality.
We report the results of fine-tuning only the final convolution layer in \cref{table:general}, as this approach yields relatively better performance.

\textbf{2. Fine-Tuning UNet with a joint image-depth VAE (LDM3D, LDM3D+)}: LDM3D trains a VAE to jointly encode and decode images and depth maps, followed by fine-tuning a UNet model in this latent space.
We hypothesize two possible scenarios depending on the VAE training:
First, if the depth input minimally influences the latent representation\textemdash meaning the latent space does not effectively encode depth information\textemdash and the decoder is trained to predict depth aligned with the image, the latent space remains largely unchanged.
While this may help prevent quality degradation, the insufficient encoding of depth in the latent limits the denoising process from learning meaningful depth features.
Alternatively, if the encoder integrates depth information and produces a disentangled latent space, this setup may resemble the +CNN experiment, potentially leading to quality degradation, as suggested by the results in \cref{appdix_fig:ablation_method}.

\textbf{3. 2-stage generation using ControlNet (ControlNet-Norm, Seg, All)}: 
We report the results of using single, and all four intrinsics (depth, normal, segmentation, line) as conditions by feeding each intrinsic into corresponding ControlNet for conditional image generation, as shown in \cref{table:general}.
For single intrinsic conditions, we report surface normal and segmentation map results since it yielded relatively better performance compared to using depth map and line drawings.
The main drawback of the 2-stage generation ($c\rightarrow i\rightarrow z$) is that $c \rightarrow i$ cannot leverage image domain information, causing inconsistencies which propagate through $x$. 
\cref{appdix_fig:ablation_method} visualizes segmentation maps generated without image information, showing that while object boundaries are well distinguished, the lack of image or other intrinsic information leads to unnatural spatial layouts and overall unrealistic scene modeling.
When these segmentation maps are used as conditions for ControlNet-based image generation, spatial inconsistencies still occur.
The qualitative degradation of intrinsic quality is also evident in the LoRA-only ablation in \cref{table:ablation}.
By contrast, in our method, since images and intrinsics are modeled complementarily, both intrinsics and images are generated with greater spatial consistency.

In summary, baseline methods either fail to incorporate meaningful information from the intrinsic domain into images, or significantly harms the quality of the base model. 

\subsection{Qualitative Results} \label{appendix:qualitative}

\begin{figure}[h]
  \centering
  \includegraphics[width=0.98\linewidth]{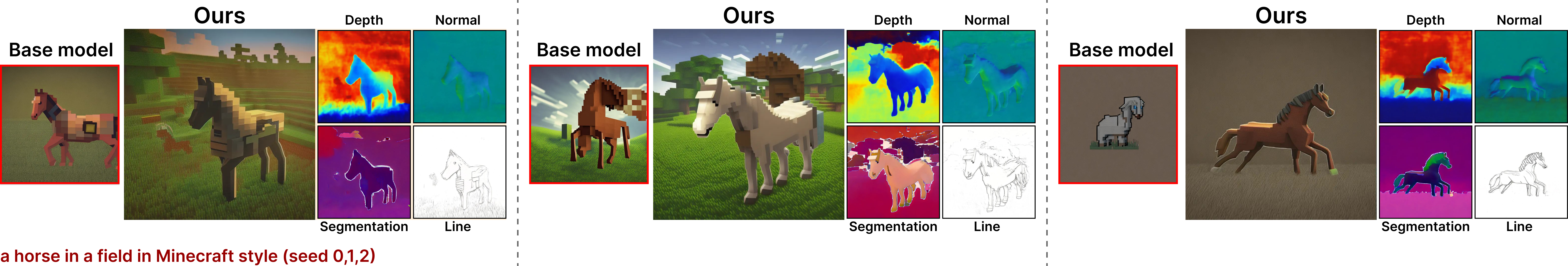}
  \caption{
  Examples of samples generated from the same prompt (a horse in a field in Minecraft style) with different seeds. 
  }
  \label{appdix_fig:diversity}
\end{figure}

\newpage
\clearpage

\begin{figure}[h]
  \centering
  \includegraphics[width=0.8\linewidth]{AnonymousSubmission/figures/sd21_parti_drop.pdf}
  \caption{
  Examples generated from Parti prompts using SD2.1 and ours with drop scheduling.
  The red box highlights spatial inconsistencies. 
  The top six rows illustrate successful examples of \model{}, while the bottom two rows illustrate failure cases.
  }
  \label{appdix_fig:sd21_parti_drop}
\end{figure}

\begin{figure}[h]
  \centering
  \includegraphics[width=0.8\linewidth]{AnonymousSubmission/figures/sd21_multi_drop.pdf}
  \caption{
  Examples generated from Multi prompts using SD2.1 and ours with drop scheduling. 
  The red box highlights spatial inconsistencies. 
  The top six rows illustrate successful examples of \model{}, while the bottom two rows illustrate failure cases.
  }
  \label{appdix_fig:sd21_multi_drop}
\end{figure}

\begin{figure}[h]
  \centering
  \includegraphics[width=0.8\linewidth]{AnonymousSubmission/figures/sd21_parti_gaussian.pdf}
  \caption{
  Examples generated from Parti prompts using SD2.1 and ours with Gaussian scheduling. 
  The red box highlights spatial inconsistencies. 
  The top six rows illustrate successful examples of \model{}, while the bottom two rows illustrate failure cases.
  }
  \label{appdix_fig:sd21_parti_gaussian}
\end{figure}

\begin{figure}[h]
  \centering
  \includegraphics[width=0.8\linewidth]{AnonymousSubmission/figures/sd21_multi_gaussian.pdf}
  \caption{
  Examples generated from Multi prompts using SD2.1 and ours with Gaussian scheduling. 
  The red box highlights spatial inconsistencies. 
  The top six rows illustrate successful examples of \model{}, while the bottom two rows illustrate failure cases.
  }
  \label{appdix_fig:sd21_multi_gaussian}
\end{figure}

\begin{figure}[h]
  \centering
  \includegraphics[width=0.8\linewidth]{AnonymousSubmission/figures/sdxl_multi.pdf}
  \caption{
  Examples generated from Multi prompts using SDXL and ours with drop scheduling (top 4 rows) and Gaussian scheduling (bottom 4 rows). 
  The red box highlights spatial inconsistencies. 
  }
  \label{appdix_fig:sdxl_multi}
\end{figure}

\begin{figure}[h]
  \centering
  \includegraphics[width=0.8\linewidth]{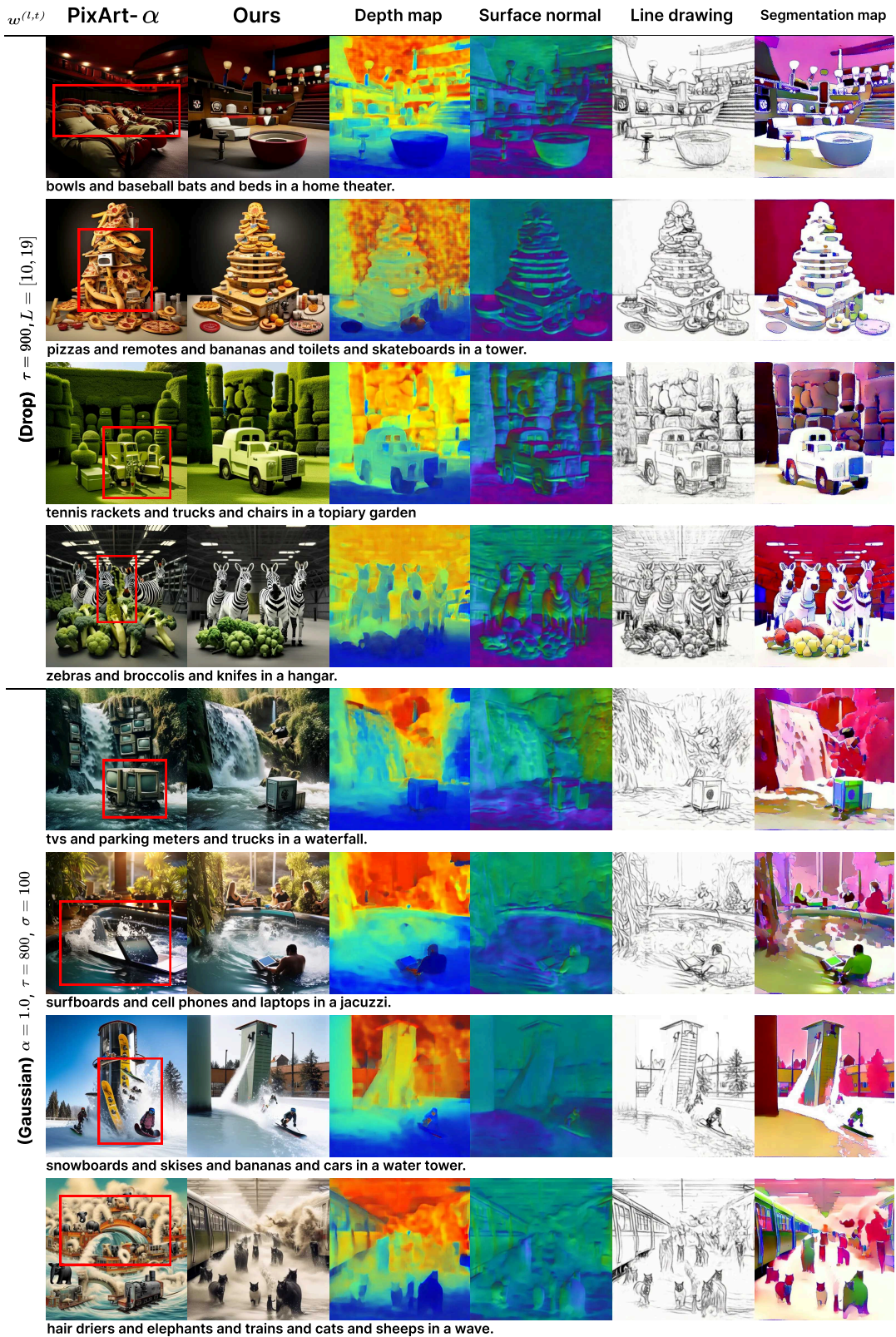}
  \caption{
  Examples generated from Multi prompts using PixArt-$\alpha$ and ours with drop scheduling (top 4 rows) and Gaussian scheduling (bottom 4 rows). 
  The red box highlights spatial inconsistencies. 
  }
  \label{appdix_fig:pixart_multi}
\end{figure}

\newpage
\clearpage

\section{InterHand Experiment Details and Results}
\label{appendix:hand}

\subsection{Experimental details}

\begin{figure}[h]
  \centering
  \includegraphics[width=0.55\linewidth]{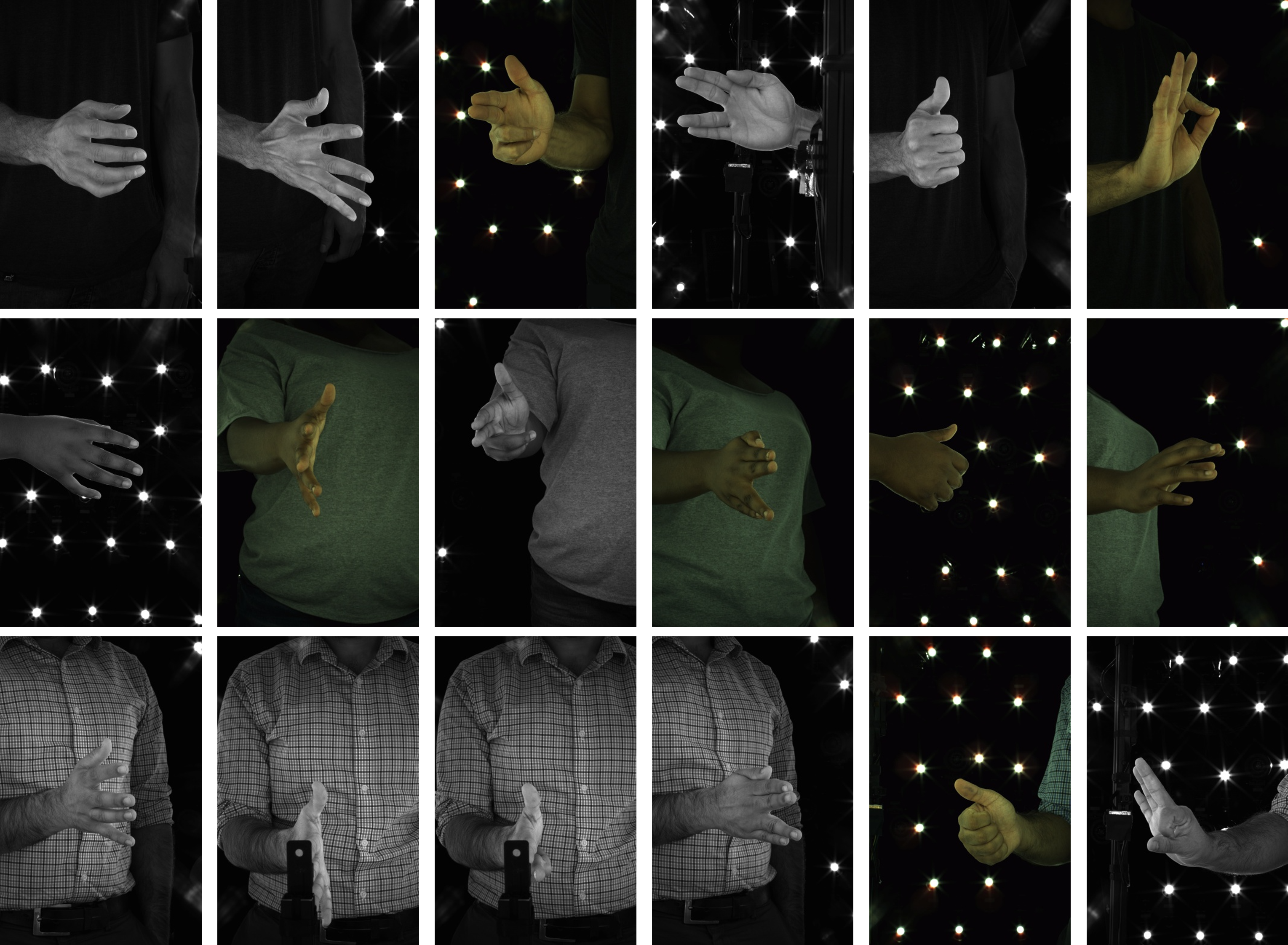}
  \caption{
  Visualization of samples from the InterHand2.6M dataset, where each row corresponds to the same recording and each column represents the same gesture.
  }
  \label{appdix_fig:hand_data}
\end{figure}

The InterHand2.6M dataset~\cite{hand} contains 2.6 million frames of single and interacting hands along with annotations, while we use only a very small subset of samples from this dataset. 
Specifically, to preserve the diversity of gestures and subjects, we select the middle frame from a randomly selected camera for each of the total 3,493 gestures contained in 36 recordings. 
We fine-tune each base model using a rank-8 LoRA applied to the attention layers, training with a batch size of 8 for 3,000 steps. The LoRA trained on the base model is then applied identically to the image domain in \model{} for evaluation.

\subsection{Qualitative results}

\begin{figure}[!h]
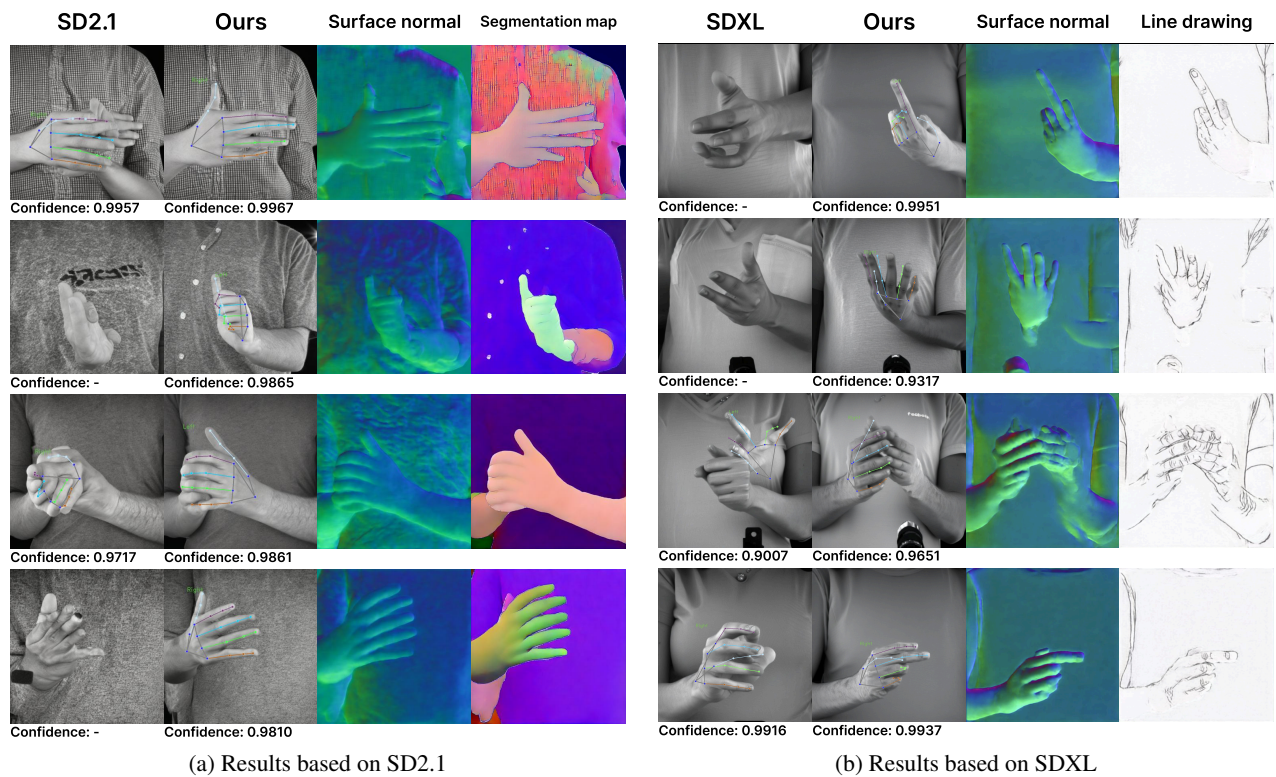

    \centering
    \subfloat[Results based on SD2.1]{           
        \includegraphics[width=0.46\linewidth]{AnonymousSubmission/figures/hand_sd21.pdf} 
    }
    \hspace{0.5em}
    \subfloat[Results based on SDXL]{           
        \includegraphics[width=0.46\linewidth]{AnonymousSubmission/figures/hand_sdxl.pdf} 
    }
    \caption{Visualizations of successful examples of \model{} in hand generation tasks.}
    \label{appdix_fig:hand}
\end{figure}

\cref{appdix_fig:hand} presents visualizations of successful examples from \model{} using Gaussian sampling. 
Both models represent the geometric structures of each finger in the surface normals, while the line drawing and segmentation map delineate the boundaries of each finger.
By incorporating these details into the images, they more clearly represent the complex structure of the hand.


\section{Qualitative Results for Ablating Intrinsics} \label{appendix:abl_intrinsic}

\begin{figure}[h]
  \centering
  \includegraphics[width=0.75\linewidth]{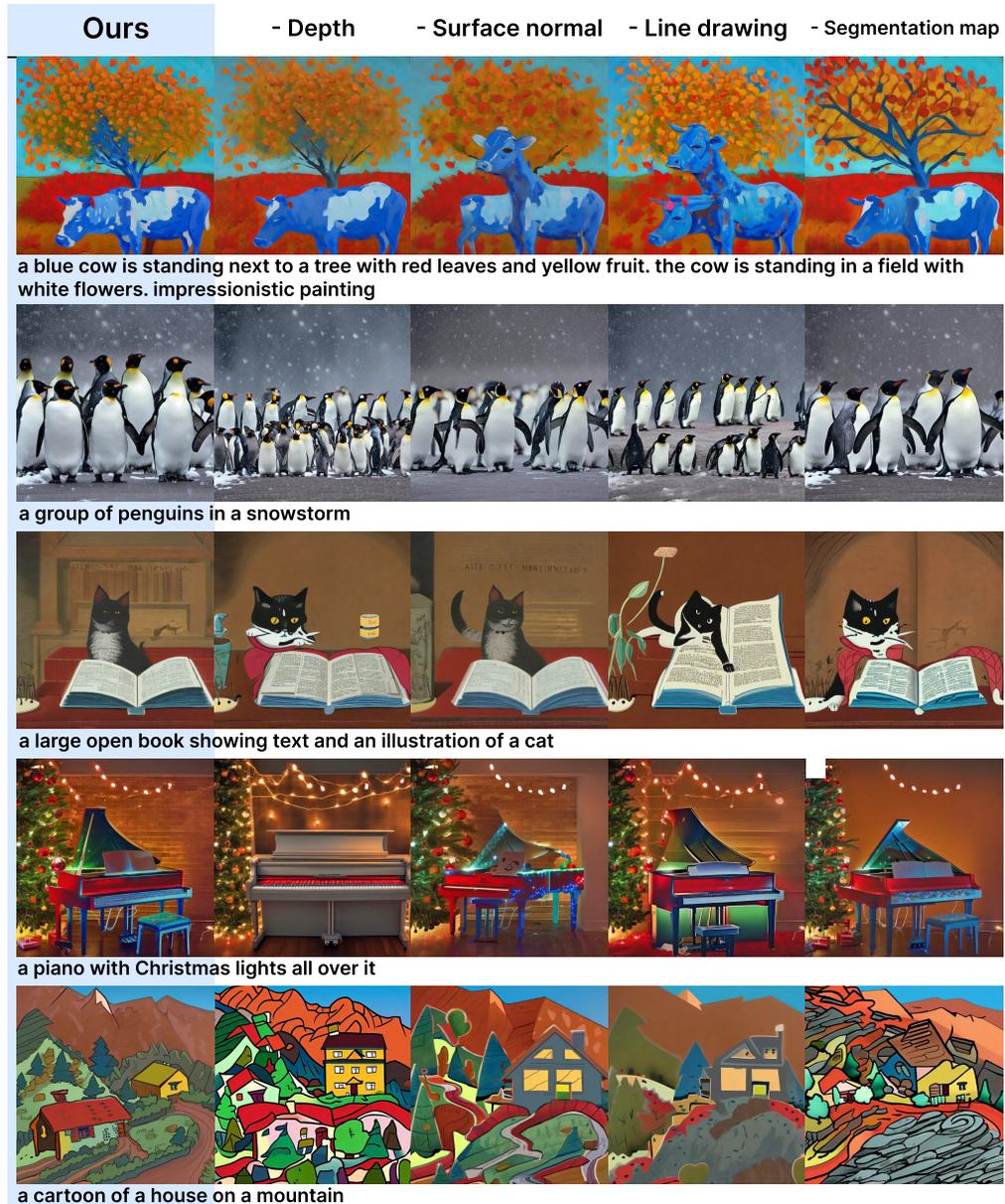}
  \caption{
  Visualization of the ablation results for each intrinsic component constituting \model{}. 
  }
  \label{appdix_fig:ablation_intr}
\end{figure}

\cref{appdix_fig:ablation_intr} visualizes the effects of each intrinsics constituting \model{}. 
When the intrinsics representing the geometric structure of the scene\textemdash depth map and surface normal\textemdash are removed, the resulting images have a reduced sense of depth and an inaccurate representation of object shapes. 
Also, when the intrinsics that convey the semantic boundaries between objects in the scene\textemdash line drawing and segmentation map\textemdash are removed, the representation of object boundaries becomes less precise, and similar textures tend to merge together.

\section{Qualitative Results and Analysis for Ablating Training Settings} \label{appendix:abl_settings}

\begin{figure}[h]
  \centering
  \includegraphics[width=0.75\linewidth]{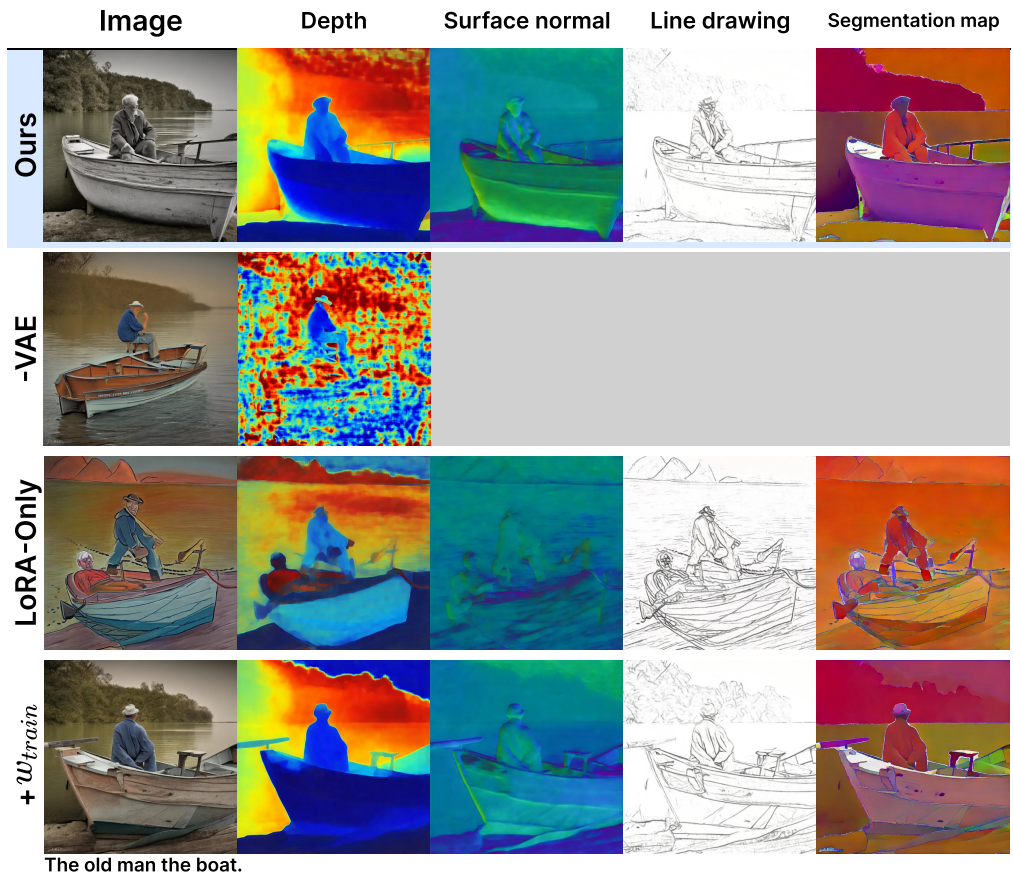}
  \setlength{\belowcaptionskip}{-2pt}  
  \caption{
  Visualization of the ablation results corresponding to \cref{table:ablation}. For the results of "-$w^{(l, t)}$" please refer to \cref{fig:schedule}.
  }
  \label{appdix_fig:ablation_hyperparam}
\end{figure}

We report the results of intrinsics and images generated under ablated training settings in \cref{table:ablation}.
"-$w^{(l, t)}$" refers to the experiment where we sample images without scheduling weights by fixing $w^{(l, t)} = 1$. 
"-VAE" denotes the experiment without intrinsic VAE, where we use a depth map encoded by a pre-trained VAE.
"LoRA-only" represents the results of training LoRA weights only from the intrinsic domain without sharing information from the image domain. 
"+$w_{train}$" refers to the experiment where we also use $w_g^{(l, t)}$ during training, and "+resolution" denotes training our model at a higher resolution of 768$\times$768.
For visualizations of "-$w^{(l, t)}$", please refer to \cref{fig:schedule}.

First, when we train the model to generate depth encoded by the pre-trained VAE instead of using an intrinsic VAE, we observe that the generation of intrinsics becomes highly unstable and of poor quality. As a result, the images fail to extract useful information from the intrinsics.
Second, when we train only the LoRA weights for the intrinsic domain without incorporating cross-domain self-attention from the image domain, the quality of the generated intrinsics degrades, which also affects the images.
Finally, applying scheduling to the image domain during training improves the quality of the generated intrinsics. However, this leads to weaker alignment between the images and the intrinsics, resulting in a stronger tendency to generate spatially inconsistent images.

\section{Limitations and Future Works}
\label{appendix:limitation}
While our work reveals the potential of incorporating intrinsics into generating images, there is still room for improvement in performance and computational efficiency.
Typically, our LoRA-based learning method, which corresponds to approximately 1\% of the denoising network parameter, can effectively preserve the image quality and diversity of the base model, but there remains potential to improve the quality and alignment of the generated intrinsic.
To address this issue, co-generating images and intrinsics within a single model trained from scratch would be a promising future direction.
Furthermore, we believe that increasing the variety of intrinsics and using training data with more complex structures will improve the effectiveness of our approach.
Finally, we expect that enriching quantitative measurements of spatial inconsistency would be a promising future work.
Also, investigating the advantages of incorporating intrinsics in various downstream tasks such as image classification \citep{lee2023learning} would be a promising direction, which could be further combined with additional data pre-processing such as data augmentations \citep{shorten2019survey,lee2022improving,hwang2022selecmix}.

\end{document}